\documentclass{article} 
\usepackage[final]{colm2026_conference}

\usepackage{microtype}
\usepackage{hyperref}
\usepackage{url}
\usepackage{booktabs}
\usepackage{graphicx} 
\usepackage{amsmath} 
\usepackage{float}
\usepackage{tabularx}
\usepackage{longtable}   
\usepackage{subcaption}
\usepackage{placeins}

\usepackage{lineno}

\definecolor{darkblue}{rgb}{0, 0, 0.5}
\hypersetup{colorlinks=true, citecolor=darkblue, linkcolor=darkblue, urlcolor=darkblue}

\title{Shared Circuits for Shared Grammar:\\Tracing Subject-Verb Agreement Across Languages}

\author{Isabella Gidi \\
Harvard University \\
\texttt{isabellagidi@college.harvard.edu}
\And
Antonio Almudévar \\
University of Zaragoza
\And
Core Francisco Park \\
Harvard University
\And
Naomi Saphra \\
Boston University
\And
Ricard Marxer \\
Univ Toulon, Aix Marseille Univ, CNRS, LIS \& ILLS
}

\usepackage{soul}
\definecolor{coreorange}{RGB}{255, 165, 0}
\sethlcolor{coreorange}

\begin{document}

\ifcolmsubmission
\linenumbers
\fi

\maketitle

\begin{abstract}
Multilingual large language models often generalize across languages, and prior work suggests that their internal mechanisms can overlap cross-lingually. It remains unclear, however, when such sharing emerges and whether it varies with the overt realization of the same grammatical operation. We investigate this question for present-tense subject-verb agreement, a morphosyntactic process that varies substantially across languages and is only weakly expressed in English. Using activation patching and attention analysis across 29 languages and five open-source model families, we identify the attention heads causally implicated in agreement and compare these head-level signatures across languages. We find that languages with overt person/number inflection exhibit more similar agreement circuitry than non-conjugating languages, with the strongest sharing appearing when the analysis isolates recovery of the inflectional contrast itself. English provides an informative bridge case, becoming more similar to conjugating languages precisely in contexts where overt agreement is required. Finally, many implicated heads display similar attention patterns across languages, suggesting that cross-lingual overlap reflects shared functional roles as well as shared localization. Together, these results indicate that multilingual LLMs reuse partially shared computational structure for morphosyntactic agreement rather than relying on fully separate language-specific solutions.
\end{abstract}

\section{Introduction}

Multilingual large language models (LLMs) generalize across a wide range of languages, yet we still do not understand the extent to which their internal computations are shared. In particular, it remains unclear when multilingual models rely on language-specific mechanisms and when they reuse common internal circuitry. This distinction matters because it shapes what kinds of cross-lingual generalization we should expect inside the model: if the same computation is implemented by shared circuitry, then interpretability findings, causal interventions, and alignment methods may transfer across languages; if it is implemented differently, then analyses and interventions developed in one language may fail to generalize to others.

This question is especially important for morphology and syntax. Languages differ in these domains in ways that may not map straightforwardly onto a single language-agnostic internal representation, especially when a high-resource language lacks the relevant feature altogether. This raises a central mechanistic question: when multilingual models process the same morphosyntactic phenomenon across typologically different languages, do they recruit \textit{shared} internal circuitry or \textit{independent} language-specific solutions? 

Prior work has established that cross-lingual circuit overlap can occur in focused case studies; the unresolved question is what determines when multilingual models reuse the same circuitry and when they recruit language-specific mechanisms. 

We investigate this question through the lens of subject-verb agreement (Appendix~\ref{app:task}). Present-tense verb conjugation provides a particularly informative testbed because the same grammatical dimensions are realized differently across languages: person and number are overtly marked in many languages, absent in languages such as Chinese, and only partially marked in English. Here, we use \textit{conjugation} to mean changes in a verb's surface form as a function of the subject's grammatical person and number. Using activation patching and attention analysis across 29 languages and six checkpoints from five model families, we test whether agreement-circuit similarity varies systematically with the requirement to overtly realize this grammatical operation, rather than asking only whether cross-lingual overlap can be detected. We further distinguish circuitry that broadly supports recovery of the context-appropriate target form from circuitry that specifically restores the inflectional contrast in controlled minimal pairs.

Our main findings are as follows:
\begin{enumerate}
    \item \textbf{Conjugation circuitry is partially shared across languages:} Among languages with overt subject-verb agreement, we find substantial overlap in the attention heads implicated in conjugation.

    \item \textbf{Shared circuitry is strongest for inflection-specific recovery:} We distinguish between broader recovery of the context-appropriate target form and stricter recovery of the inflectional contrast itself. Similarity is highest under the minimal-pair metric, suggesting that multilingual models share circuitry most strongly for overt inflectional agreement, while broader target-form recovery is only partially shared and more variable across languages.

    \item \textbf{Shared circuitry tracks shared grammatical structure:} Languages with little or no person-based verb inflection show substantially lower similarity to these head-level signatures. English provides an informative bridge case: it differs from strongly conjugating languages overall, but its third-person singular cases align more closely with the shared conjugation circuitry than its non-inflecting cases do.

    \item \textbf{Overlapping heads play similar functional roles:} Across conjugating languages, shared heads show similar attention patterns, suggesting that cross-lingual overlap reflects shared information routing rather than only shared localization.
\end{enumerate}

\section{Related Work}

\paragraph{Shared and language-specific multilingual mechanisms.}
A substantial literature suggests that multilingual models often rely on partially shared cross-lingual structure, especially for semantics, factual recall, and safety-related behavior \citep{wendler-etal-2024-llamas, dumas-etal-2025-separating, NEURIPS2025_2e94772e}. Work on grammar and morphosyntax likewise finds evidence of shared organization, including overlapping grammatical subspaces, shared neuron sets, and agreement-related components \citep{stanczak-etal-2022-neurons, mueller2022agreement_neurons, ferrando-costa-jussa-2024-similarity}. At the same time, other studies identify language-specific neurons or features and show that cross-lingual transfer and interventions do not generalize uniformly across languages \citep{tang2024neurons_acl, deng-etal-2025-unveiling, zhang2025the}. Taken together, this literature suggests that the key question is not whether multilingual computation is globally shared or language-specific, but \emph{when} each pattern arises. This question is especially important for morphosyntax, where languages may share broad meaning while differing in whether they overtly realize a grammatical operation at all. We therefore study present-tense subject-verb agreement as a test case for when multilingual models reuse shared internal mechanisms and when they diverge.

\paragraph{From shared representations to shared circuitry.}
Cross-lingual structure has been studied with probing, representational analyses, neuron and feature methods, and causal interventions \citep{mikhailov-etal-2021-morph, brinkmann-etal-2025-large, jing-etal-2025-lingualens, mueller2022agreement_neurons}. Our work is closest to causal studies of agreement and cross-lingual circuit similarity \citep{mueller2022agreement_neurons, ferrando-costa-jussa-2024-similarity, zhang2025the}, but differs in three ways most relevant to our question. First, we use tightly controlled multilingual minimal pairs that isolate the inflectional contrast. Second, we compare a broader and more typologically varied set of languages, including languages with rich agreement, partial agreement, and no overt agreement in the tested contrasts. Third, we use two complementary recovery metrics: target-logit recovery, which measures broader recovery of the context-appropriate form under a subject change, and minimal-pair recovery, which measures recovery of the inflectional contrast itself. This allows us to ask separately whether multilingual models share mechanisms for broader subject-conditioned prediction and whether they share mechanisms for implementing overt agreement morphology.

\section{Experimental Setup}
\label{sec:experimental_setup}

\paragraph{Models evaluated.} We investigate conjugation circuitry across five open-source model families: BLOOM \citep{scao2023bloom}, Gemma \citep{riviere2024gemma2}, Llama \citep{touvron2023llama2, grattafiori2024llama3herdmodels}, Mistral \citep{jiang2023mistral7b}, and Qwen \citep{yang2024qwen2}. The full model list is provided in Appendix~\ref{app:modelss}, with additional analyses of model size and instruction tuning in Appendix~\ref{app:robustness_checks}.

\paragraph{Verb dataset.}
We construct our candidate verb inventory from structured lexical and morphological resources. For 33 languages, we extract verb infinitives and their corresponding present-tense inflected forms from UniMorph \citep{batsuren-etal-2022-unimorph}, which provides isolated morphological paradigms. We additionally use the Chinese Verb Semantic Feature Dataset \citep{deng2024cvfd} to obtain verb lemmas for Chinese, providing a high-resource non-conjugating comparison language. At this stage, these resources provide only the lexical items and their relevant forms; the experimental prompts are subsequently constructed from these entries using the controlled template described below. Across languages, the number of distinct verb infinitives with corresponding conjugated forms ranges from 56 to 23,782, with an average of 1,750.74. In total, our initial inventory spans 34 languages.

\paragraph{Inflection contrasts.} To keep the analysis tractable while covering the main morphosyntactic dimensions of interest, we evaluate a representative subset of bidirectional person and number contrasts. Specifically, we consider shifts in person while holding number constant ($1\text{sg} \leftrightarrow 2\text{sg}$, $1\text{sg} \leftrightarrow 3\text{sg}$, and $1\text{pl} \leftrightarrow 3\text{pl}$), as well as a shift in number while holding person constant ($3\text{sg} \leftrightarrow 3\text{pl}$). Of these four pairs, English has an inflectional change on two ($1\text{sg} \leftrightarrow 3\text{sg}$, $3\text{sg} \leftrightarrow 3\text{pl}$).

\paragraph{Prompt template.} For each language, we generate zero-shot prompts designed to elicit the target inflection. By using a common prompt structure, we largely isolate differences in agreement-related circuitry. The prompts follow a common template of the form: \textit{``Conjugation of the verb [infinitive] in present tense: [pronoun] [conjugation]''}, adapted into each target language (Adaptations in Appendix~\ref{app:prompts}).  This deliberately structured elicitation format preserves positional alignment between clean and corrupted prompts, which is important for controlled activation patching, but trades naturalism for experimental control. In preliminary experiments across models, this phrasing maximized the probability assigned to the correct conjugation.

\paragraph{Tokenization filtering.}
\label{paragraph:tokenization1}
To construct tightly controlled minimal pairs in conjugating languages, we apply a model-specific tokenization filter and retain only verbs whose clean and corrupted conjugated forms differ only in the final token. This preserves alignment in the prefix sequence and isolates the person or number contrast to the token the model must predict. Examples of retained and excluded cases are provided in Appendix~\ref{app:tokenization_examples}.

\paragraph{Behavioral filtering.} Because our goal is to analyze the mechanisms involved in successful agreement computation, we also remove examples where the model does not produce the desired verb inflection. Following prior work on isolating task-relevant circuitry \citep[e.g.,][]{todd2024function_vectors}, this restriction focuses the analysis on successful realizations of grammatical agreement rather than heterogeneous error behavior. We thus interpret our analyses as characterizing the circuitry used in successful conjugation behavior rather than overall language competence.

\paragraph{Language-model coverage.} 
Official language support varies across model families, but many models achieve sufficient accuracy on languages beyond those explicitly documented in their training descriptions. We include a language-model combination in our analysis only if at least 50 prompts remain after tokenization and behavioral filtering. Starting from 34 languages, this yields coverage for 29 languages in the main analyses. These retained languages include 24 languages with overt agreement patterns, 4 languages with no overt agreement in the tested contrasts, and English as an intermediate case, since it marks agreement only in the third-person singular. The resulting language-model coverage and full language inventory are provided in Appendix~\ref{subsec:lang_model}.

\paragraph{Final dataset.}
After tokenization and behavioral filtering, the number of surviving prompts ranges from 96 to 2,400 across the retained language-model combinations, with an average of 1,509.59. From each retained set, we evaluate 50 prompts for activation patching.

\section{Analytic Methods}
\label{subsec:mechanistic_techniques}

To identify and characterize the internal circuitry involved in subject-verb agreement, we use attention-output patching to localize causally important attention heads and attention analysis to characterize their behavior.

\subsection{Attention-output Patching}

\begin{figure}[htbp]
    \centering
    \includegraphics[width=0.90\textwidth]{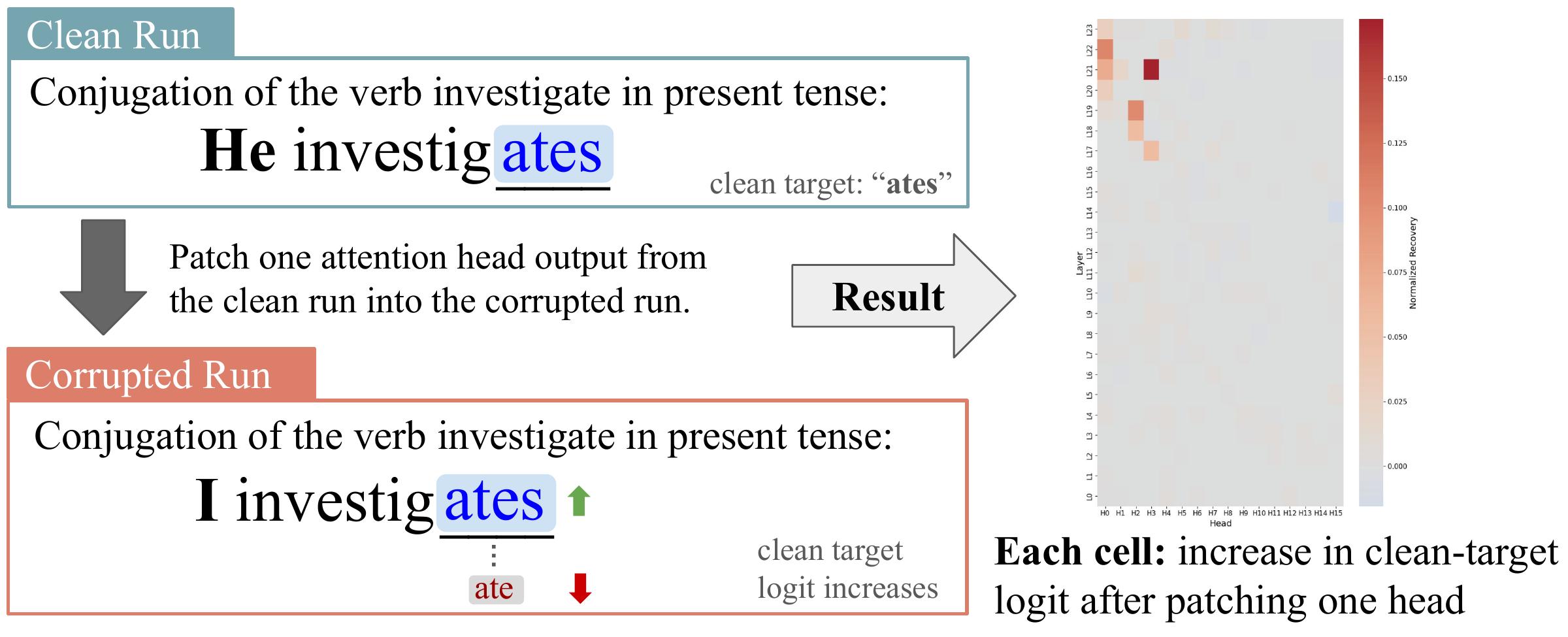} 
    \caption{\textbf{Example of activation patching.}
    The clean and corrupted prompts differ only in the subject pronoun and target inflection, with the contrast isolated to the final token. For each layer-head pair, we patch that head's output from the clean run into the corrupted run and measure the resulting increase in the clean-target logit. The resulting heatmap summarizes the causal contribution of each head.}
    \label{fig:activation_patching_example}
\end{figure}

We use attention-output patching \citep{vig2020investigating, zhang2024activationpatching} to identify the attention heads causally important for predicting the target verb form under subject manipulations (Figure~\ref{fig:activation_patching_example}).

\paragraph{Minimal pairs.} Minimal pairs are constructed as described in Section~\ref{sec:experimental_setup}. For each pair, we define a \emph{clean} prompt corresponding to the target inflection and a \emph{corrupted} prompt corresponding to the contrastive inflection. For example, when evaluating a $1\text{sg} \rightarrow 3\text{sg}$ shift, the third-person form is the clean prompt and the first-person form is the corrupted prompt.

\paragraph{Head-level patching.} During a forward pass on the corrupted prompts, we replace the output of attention head $h$ in layer $\ell$ with its cached activation from the clean run at the corresponding sequence position. Let $\mathbf{L}_{\text{patch}}^{(\ell,h,i)}$ denote the resulting final-layer logit vector for prompt $i$, let $\mathbf{L}_{\text{corr}}^{(i)}$ denote the unpatched corrupted logits, and let $y_{\text{clean}}^{(i)}$ denote the clean target token.

\paragraph{Recovery metrics.} We use two complementary recovery metrics: target-logit recovery and minimal-pair recovery. We use target-logit recovery for analyses that include non-conjugating languages, and minimal-pair recovery for analyses restricted to conjugating languages. For both metrics, larger absolute values indicate stronger effects overall, with positive values reflecting promotion of the clean target token and negative values reflecting inhibition. Unless otherwise noted, each metric is computed over 50 prompts per conjugation pair, evaluated in both patching directions (e.g., $1\text{sg} \rightarrow 3\text{sg}$ and $3\text{sg} \rightarrow 1\text{sg}$), and aggregated into layer-head heatmaps.

\paragraph{Target-logit recovery.} Target-logit recovery measures the increase in the clean target token's logit relative to the corrupted baseline:
\[
M_{\text{target}}(\ell,h)=\frac{1}{n}\sum_{i=1}^n\left(\left[\mathbf{L}_{\text{patch}}^{(\ell,h,i)}\right]_{y_{\text{clean}}^{(i)}}-\left[\mathbf{L}_{\text{corr}}^{(i)}\right]_{y_{\text{clean}}^{(i)}}\right).
\]
Because this metric does not require a contrast between two different inflected forms, it can be applied both to conjugating languages and to languages or conjugation pairs without overt present-tense agreement inflection. In conjugating languages, it captures how much patching restores the clean inflected form relative to the corrupted prompt. In languages or conjugation pairs without overt agreement inflection, the clean and corrupted prompts may share the same final target token, but the subject manipulation can still change the logit assigned to that token. In those cases, the metric asks whether patching restores the clean-context preference for the shared target form. We therefore interpret the identified heads as supporting successful prediction of the target form in context, not always as encoding overt inflectional contrast.

\paragraph{Minimal-pair recovery.}
Minimal-pair recovery measures how much patching restores the clean-versus-corrupted logit contrast. For each example $i$, let $y_{\mathrm{clean}}^{(i)}$ and $y_{\mathrm{corr}}^{(i)}$ denote the competing final inflection tokens, and define the contrast for a final-layer logit vector $\mathbf{L}^{(i)}$ as
\[
\Delta^{(i)}(\mathbf{L}^{(i)})
=
L^{(i)}_{y_{\mathrm{clean}}^{(i)}}
-
L^{(i)}_{y_{\mathrm{corr}}^{(i)}}.
\]
We write $\Delta_{\mathrm{clean}}$, $\Delta_{\mathrm{corr}}$, and $\Delta(\mathbf{L}_{\mathrm{patch}}^{(\ell,h)})$ for this contrast averaged across examples under the unpatched clean run, the unpatched corrupted run, and the run in which attention head $h$ in layer $\ell$ is patched, respectively. Minimal-pair recovery is then defined as
\[
M_{\mathrm{pair}}(\ell,h)
=
\frac{
\Delta(\mathbf{L}_{\mathrm{patch}}^{(\ell,h)})
-
\Delta_{\mathrm{corr}}
}{
\Delta_{\mathrm{clean}}
-
\Delta_{\mathrm{corr}}
+
\epsilon
},
\]
where $\epsilon$ is a small constant added for numerical stability. This metric is applicable only when the clean and corrupted prompts have different final inflection tokens and therefore provides a targeted measure of how strongly patching restores the preference for the context-appropriate inflected form over its contrastive alternative.

\paragraph{Cross-lingual comparison.} To test whether languages rely on similarly localized heads for the conjugation task, we compare activation-patching heatmaps across languages, treating these heatmaps as signatures of the circuitry implicated in agreement. For each language, we flatten the layer--head heatmap into a vector and compute Pearson correlation between language pairs within the same model, conjugation pair, and patching direction. Because our goal is to compare the locations of task-relevant heads rather than the direction of their effects, our primary analyses use the absolute values of the heatmaps before flattening. Under this measure, high similarity indicates that two languages assign relatively high importance to the same layer-head locations. Across all boxplots, $n$ denotes the number of observations contributing to the plotted distribution. We report alternative similarity measures, including Spearman correlation and top-$k$ head overlap, in Appendix~\ref{app:robustness_checks_metric}.

\paragraph{Cross-model comparison.} We never compare attention-head coordinates or patch activations directly across different architectures or checkpoints. For each checkpoint, we compute cross-lingual similarity between language heatmaps matched by conjugation contrast and patching direction. We then report these within-model similarities separately by checkpoint or aggregate their scalar similarity values across checkpoints for summary analyses.

\subsection{Attention Analysis}

Activation patching identifies which heads are causally important for agreement, but not how those heads behave once implicated. To characterize their functional behavior, we analyze attention patterns \citep{li2016visualizing}, focusing on attention emitted from the final prefix position, i.e., when the model is predicting the final inflection token.

\paragraph{Top-head selection.}
For each model, language, conjugation pair, patching direction, and recovery metric, we rank attention heads by the absolute magnitude of their patching score and retain the top $k=20$ heads. We compute the attention-role statistics described below separately for each selected head. For cross-lingual comparisons, we restrict the analysis to overlapping heads: heads that appear among the top 20 for both languages within the same model, conjugation pair, patching direction, and metric. We then compare the corresponding attention-role vectors for these shared heads across the two languages.

\paragraph{Attention-role summaries.} To interpret these patterns, we partition each prompt prefix into linguistically relevant categories using tokenizer-specific alignments derived from the prompt annotations. For each example, we measure the fraction of attention mass from the final prefix position directed to the subject pronoun, the explicit infinitive span, and all remaining tokens. These role-mass values are averaged first across examples and then across patching directions within each language-pair condition. We use these summaries to compare conjugating and non-conjugating languages, as well as English non-inflecting and third-person singular contexts.

\paragraph{Cross-lingual comparison.} To assess cross-lingual reuse beyond role-mass summaries, we compare the attention-role vectors associated with heads identified across languages. In the main analysis, we evaluate whether overlapping top-ranked heads exhibit similar role profiles across languages. Together, these analyses test whether heads that are similarly localized by activation patching also exhibit similar functional behavior across languages.

\section{Results}
\label{sec:results}

\paragraph{Results roadmap.}
We organize our results by moving from structural localization to mechanistic behavior. To localize the relevant circuitry, we rely on the distinction between our two recovery metrics. The \emph{minimal-pair} metric provides a sharp test of shared \emph{inflection-specific} circuitry among conjugating languages, whereas the broader \emph{target-logit} metric allows us to test whether subject-conditioned prediction generalizes to non-conjugating languages. Finally, after identifying these cross-lingual networks, we analyze their fine-grained attention patterns to determine whether these structurally shared heads execute the same information-routing mechanisms across languages.

\subsection{Inflection-specific sharing is strongest under the minimal-pair metric}
\label{subsec:results_minimal_pair}

We first investigate whether languages with overt present-tense agreement recruit similar head-level circuitry, and how the choice of recovery metric influences this shared structure.

\begin{figure*}[htb!]
    \centering
    \begin{subfigure}[t]{0.53\textwidth}
        \centering
        \includegraphics[width=\linewidth]{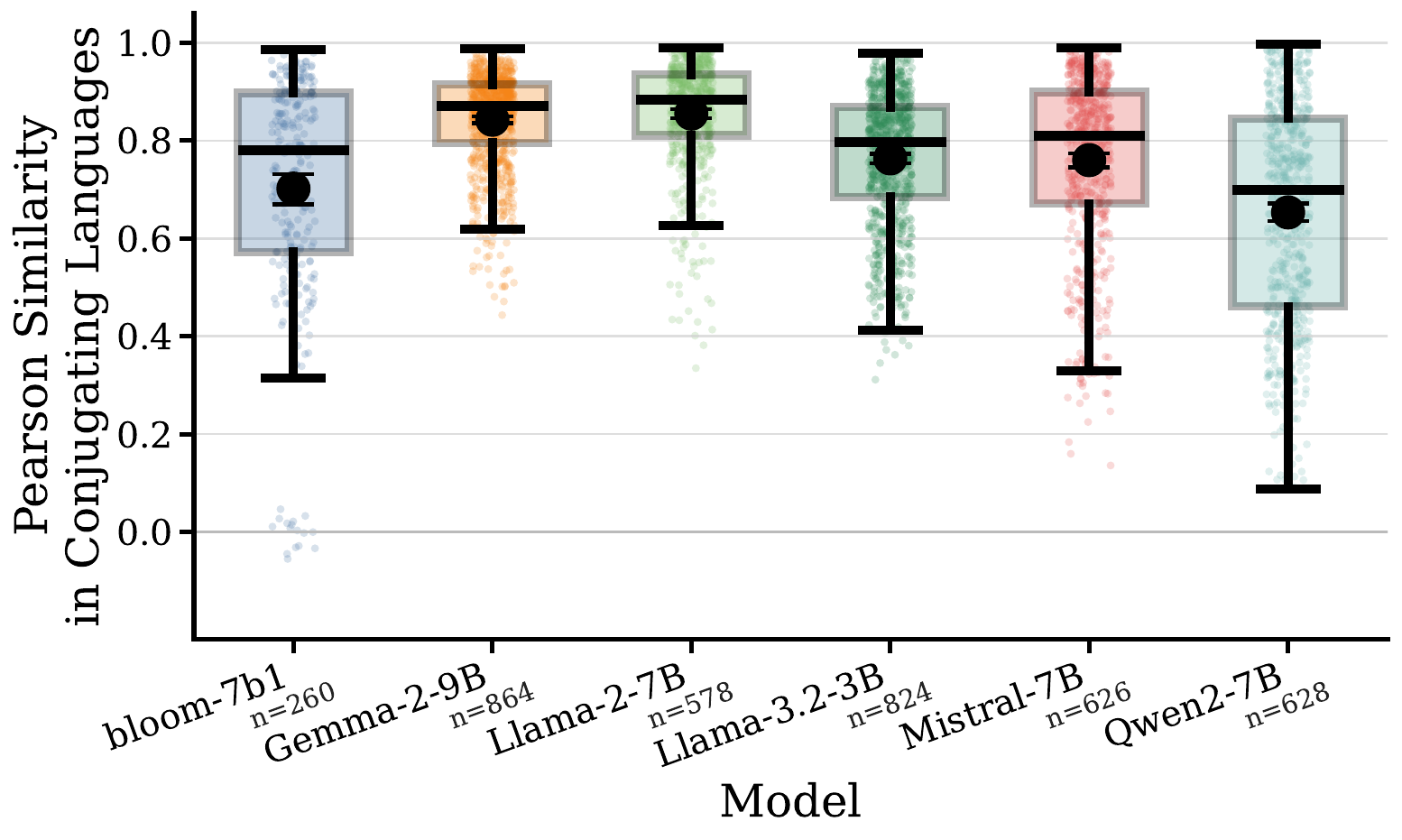}
    \end{subfigure}
    \hfill
    \begin{subfigure}[t]{0.45\textwidth}
        \centering
        \includegraphics[width=\linewidth]{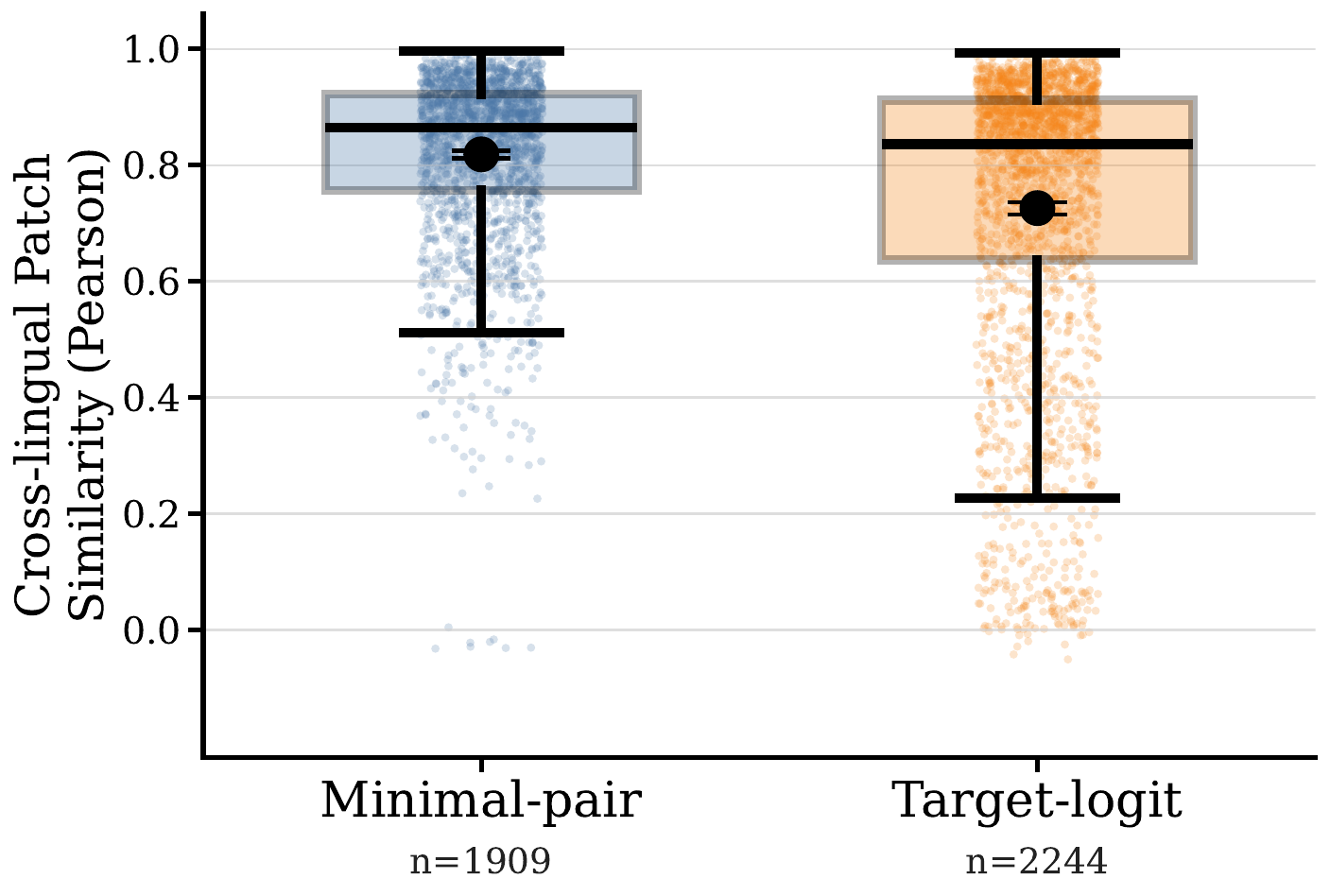}
    \end{subfigure}
    \caption{\textbf{Inflection-specific sharing is stronger under the minimal-pair metric.}
    \textbf{Left:} Pairwise similarity among conjugating languages under the minimal-pair metric, shown separately by model checkpoint. Similarity is broadly positive across families, indicating shared inflection-related circuitry across languages with overt agreement.
    \textbf{Right:} Pairwise similarity among conjugating languages under the minimal-pair and target-logit metrics. Similarity is higher and less variable under the minimal-pair metric, indicating that cross-lingual sharing is strongest when the analysis isolates recovery of the inflectional contrast itself.}
    \label{fig:conjugating_metric_comparison}
\end{figure*}

\paragraph{Conjugating languages share robust inflectional circuitry.}
Under the minimal-pair metric, pairwise similarities among conjugating languages are strongly positive across model families (Figure~\ref{fig:conjugating_metric_comparison}, left). This indicates that the head-level structures implicated by activation patching are broadly shared across inflecting languages. While the exact strength and dispersion of this pattern vary by model architecture, language-level similarity matrices show that the effect is broadly distributed across languages rather than being restricted to a small set of closely related language pairs (Appendix~\ref{app:heatmaps}). This positive cross-lingual similarity also persists across all individual conjugation contrasts and patching directions (Appendix~\ref{app:direction-split}).

\paragraph{Sharing peaks during strict inflection-specific recovery.}
Comparing the two metrics reveals a critical functional distinction (Figure~\ref{fig:conjugating_metric_comparison}, right). Similarity among conjugating languages is higher and less variable under the minimal-pair metric than under the target-logit metric. This demonstrates that multilingual models share circuitry most intensely when restoring the strict inflectional contrast itself. Conversely, the broader prediction of the context-appropriate target form (measured by target-logit) exhibits only partial cross-lingual sharing, indicating it is more sensitive to language-specific computation.

\paragraph{Summary.}
Collectively, these results support a graded view of cross-lingual structural alignment. Rather than relying on entirely disjoint, language-specific mechanisms for present-tense agreement, models reuse a shared foundational circuitry. Crucially, this sharing is most pronounced in the specific circuits dedicated to resolving the inflectional contrast, whereas broader target-form prediction introduces more language-specific variance. Auxiliary analyses regarding model size, language-specific task capability, instruction tuning, and BLOOM training checkpoints are detailed in Appendix~\ref{app:robustness_checks}.

\subsection{Broader target-form recovery is less shared, but still separates conjugating from non-conjugating languages}
\label{subsec:results_target_logit}

We next ask whether the broader \emph{target-logit} metric still reveals linguistically structured sharing. Because this metric can be applied even when a language lacks overt present-tense agreement, it allows us to test whether subject-conditioned target recovery generalizes to non-conjugating languages. English is treated separately, as it occupies an intermediate position by marking agreement only in the third-person singular.

\begin{figure*}[htb!]
    \centering
    \begin{subfigure}[t]{0.62\textwidth}
        \centering
        \includegraphics[width=\linewidth]{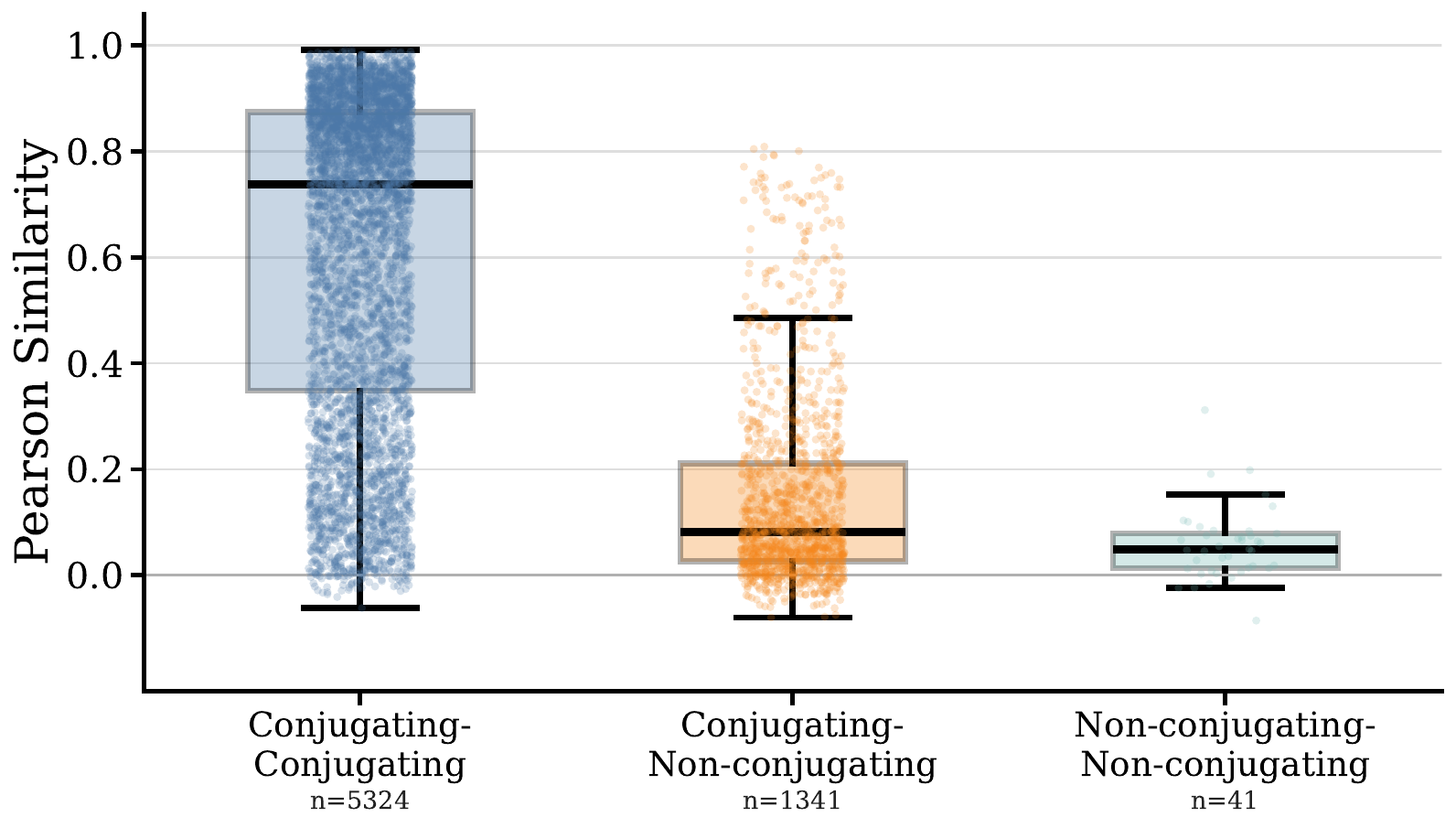}
    \end{subfigure}
    \hfill
    \begin{subfigure}[t]{0.36\textwidth}
        \centering
        \includegraphics[width=\linewidth]{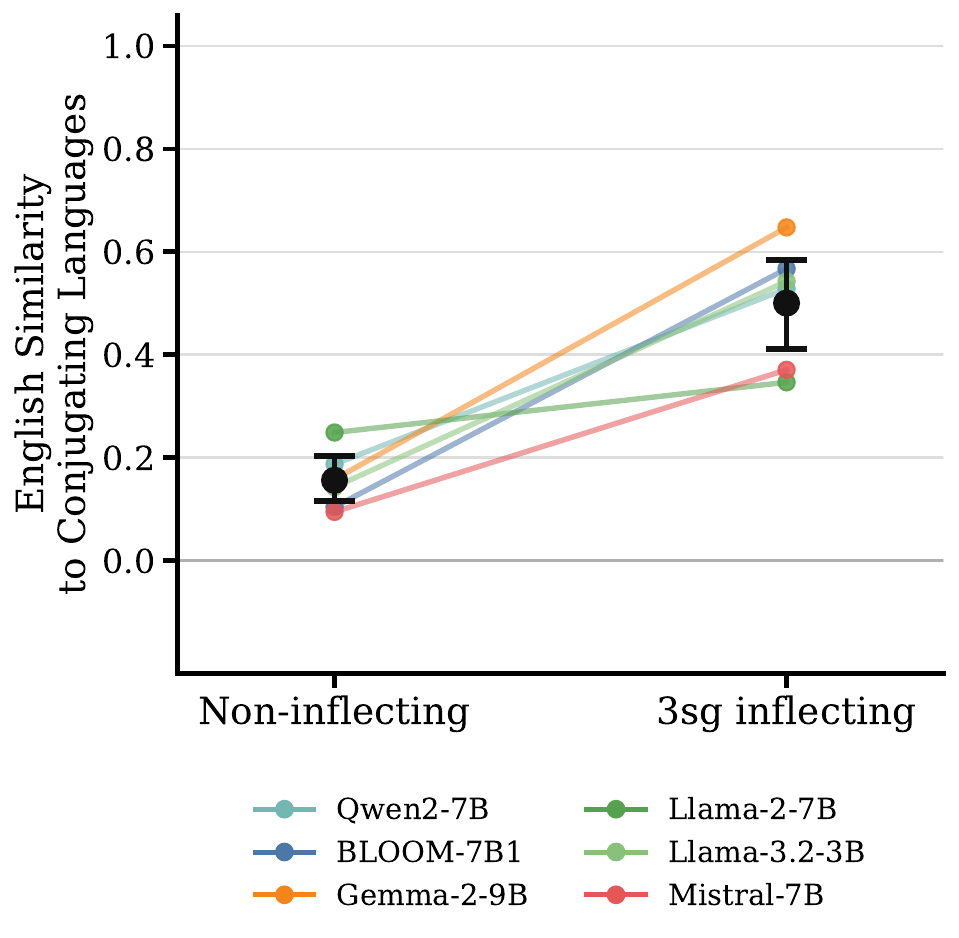}
    \end{subfigure}
    \caption{\textbf{Broader target-form recovery still tracks overt agreement structure.}
    \textbf{Left:} Pairwise similarity between activation-patching heatmaps grouped by language comparison type under the target-logit metric. Each point is a comparison between two heatmaps matched for model, conjugation pair, and patching direction. Conjugating--conjugating comparisons are substantially more similar than conjugating--non-conjugating comparisons, showing that broader target-form recovery remains structured by whether the language overtly marks agreement.
    \textbf{Right:} English similarity to conjugating languages in non-inflecting versus 3sg-inflecting contexts. Each colored line corresponds to one model and connects its mean English--to--conjugating similarity across the two condition types. Black points and error bars show the across-model mean and bootstrap confidence interval for each condition. English becomes more similar to conjugating languages precisely in the contexts where overt agreement is required.}
    \label{fig:target_logit_cross_group}
\end{figure*}

\paragraph{Language type dictates shared circuitry.}
Figure~\ref{fig:target_logit_cross_group} (left) demonstrates a clear separation by language type. Comparisons between two conjugating languages yield much higher similarity than comparisons across the conjugating/non-conjugating boundary. This indicates that the recovered head-level signatures are not generic; they are strongly shaped by whether the language overtly realizes the morphosyntactic feature. Furthermore, similarity among non-conjugating languages remains relatively low, proving that the metric is not simply recovering a universal, baseline prediction pattern for all non-conjugating languages.

This result complements Section~\ref{subsec:results_minimal_pair}: while the minimal-pair metric isolates strictly \emph{inflection-specific} shared circuitry, the broader target-logit metric still sharply delineates languages with overt agreement from those without. Despite its higher variance, it reveals a clear, cross-lingual grammatical organization.

\paragraph{English dynamically shifts based on grammatical context.}
English provides an exceptionally informative test case. Because it marks present-tense agreement only in third-person singular (3sg) contexts, it allows us to isolate whether cross-lingual similarity is driven by static language identity or by the contextual demand for an agreement computation. As shown in Figure~\ref{fig:target_logit_cross_group} (right), English exhibits a consistent upward shift from non-inflecting to 3sg-inflecting contexts. It becomes markedly more similar to conjugating languages exactly when overt agreement is required.

\paragraph{Summary.}
Taken together, these results demonstrate that while broader target-form recovery is less uniformly shared than inflection-specific recovery, it remains strictly organized by grammatical structure. Languages with overt agreement form a distinct similarity cluster, and English shifts toward this cluster precisely when agreement must be realized. Ultimately, cross-lingual sharing is driven not by language identity, but by whether the task demands the same underlying grammatical operation.

\subsection{Shared heads show agreement-focused and cross-lingually similar routing}
\label{subsec:results_attention}

Activation patching identifies which heads are causally important for agreement, but not how those heads behave once implicated. We therefore analyze the attention patterns of top-ranked heads to ask two mechanistic questions: whether routing differs across metric and language groups, and whether agreement-related routing is preserved across languages.

\paragraph{Attention profiles clearly separate conjugating and non-conjugating languages.}
We summarize the attention mass directed from the final prefix position toward the pronoun, the infinitive, and other prompt tokens across three groups: heads identified by the minimal-pair metric in conjugating languages, and heads identified by the target-logit metric in both conjugating and non-conjugating languages.

\begin{figure*}[htb!]
    \centering
    \begin{subfigure}[t]{0.52\textwidth}
        \centering
        \includegraphics[width=\linewidth]{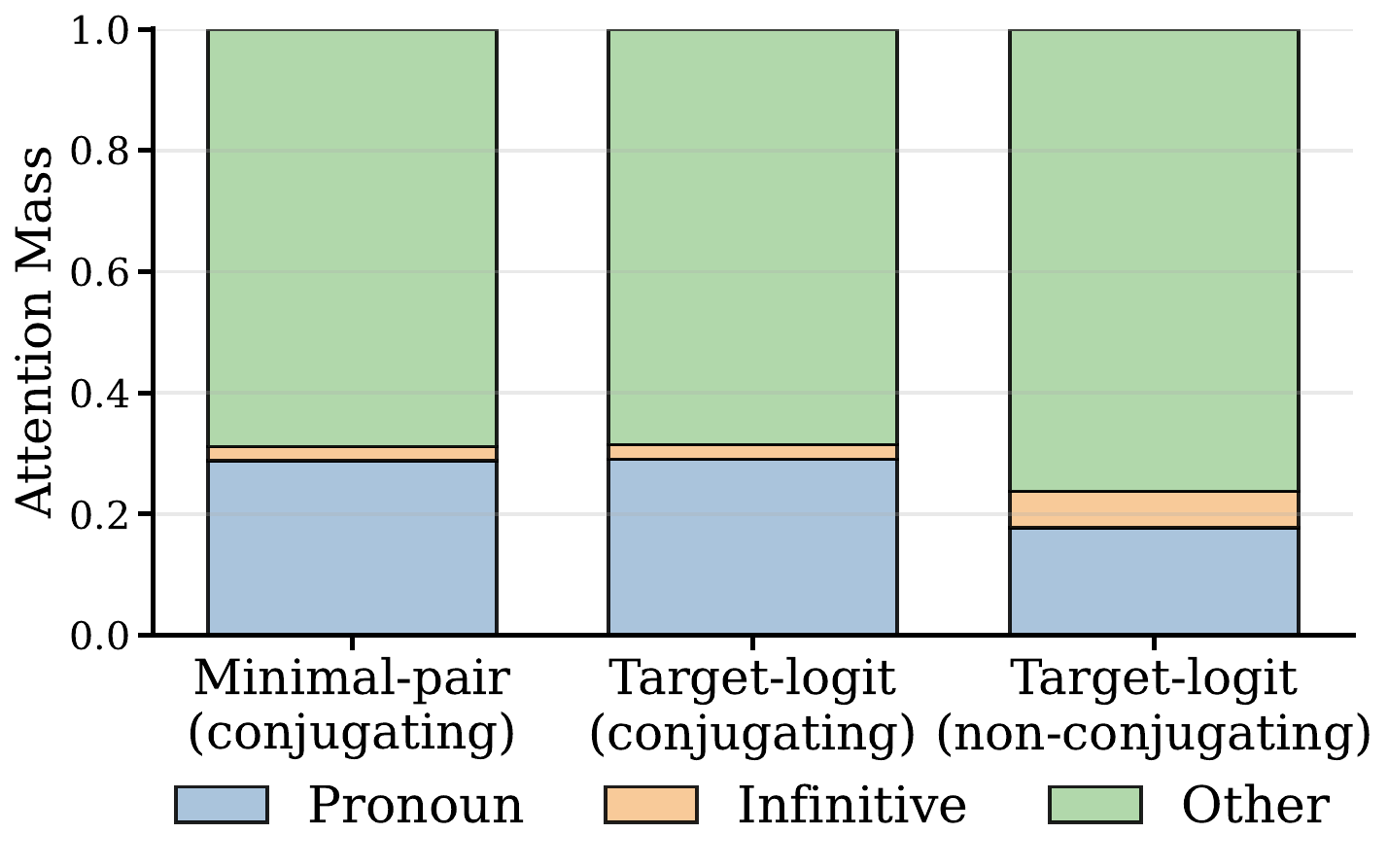}
    \end{subfigure}
    \hfill
    \begin{subfigure}[t]{0.46\textwidth}
        \centering
        \includegraphics[width=\linewidth]{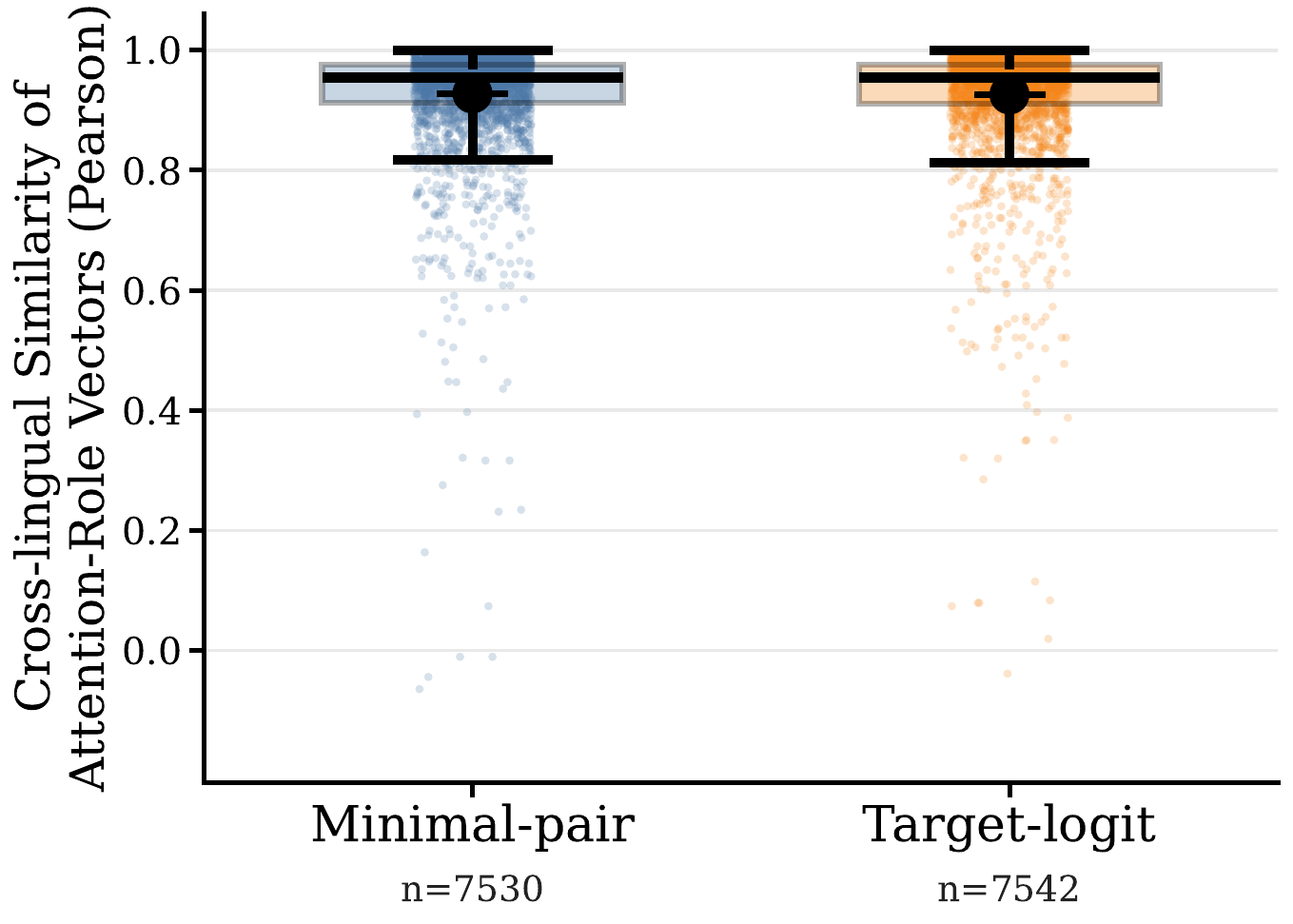}
    \end{subfigure}
    \caption{\textbf{Shared heads are agreement-focused and functionally similar across languages.}
    \textbf{Left:} Coarse attention-role composition of top heads across three groups: minimal-pair heads in conjugating languages, target-logit heads in conjugating languages, and target-logit heads in non-conjugating languages. The two conjugating groups show similar coarse routing profiles, whereas target-logit heads in non-conjugating languages allocate less attention to pronouns and more to other prompt material.
    \textbf{Right:} Cross-lingual similarity of attention-role vectors for overlapping top-20 heads, shown separately by metric. High similarity under both metrics indicates that implicated heads tend to preserve agreement-related routing roles across languages.}
    \label{fig:agreement_structure}
\end{figure*}

As shown in Figure~\ref{fig:agreement_structure} (left), the target-logit metric reveals a stark contrast between language types. In non-conjugating languages, target-logit heads direct less attention to pronoun tokens ($0.177$) and more to other prompt material ($0.762$). Conversely, both conjugating groups behave almost identically, placing greater emphasis on the pronoun (minimal-pair: $0.288$; target-logit: $0.291$) and less on other tokens (minimal-pair: $0.689$; target-logit: $0.685$). 

This functional similarity within conjugating languages provides critical context for the patching results in Section~\ref{subsec:results_minimal_pair}. Although the target-logit metric surfaces more language-specific head locations than the minimal-pair metric, the implicated heads perform the same broad routing functions. Ultimately, the choice of metric affects how strongly the recovered circuitry is shared across languages, but the fundamental information-routing behavior required for conjugation remains consistent.

\paragraph{Shared heads exhibit similar attention-role profiles across languages.}
We next ask whether high-impact heads shared across two languages exhibit similar functional profiles in both languages. As shown in Figure~\ref{fig:agreement_structure} (right), cross-lingual role similarity is exceptionally high under both metrics (mean similarity: minimal-pair $\approx 0.928$, target-logit $\approx 0.926$, based on thousands of directional language-pair observations). This strong cross-lingual functional consistency strengthens the mechanistic interpretation of our patching results. It demonstrates that cross-lingual overlap is not merely a coincidence of layer--head localization, it is consistent with partially shared information-routing roles. Furthermore, even though the two metrics isolate slightly different networks of relevant heads, the specific routing patterns within those networks remain highly consistent across conjugating languages.

\section{Discussion}

We investigated whether multilingual language models implement present-tense subject--verb agreement through shared or language-specific internal circuitry. Across 29 languages and five model families, we found evidence for a middle-ground view: multilingual models often reuse partially shared circuitry for agreement, but this sharing is neither uniform nor fully language-agnostic. Languages with overt agreement morphology exhibit substantially more similar head-level signatures than comparisons involving non-conjugating languages, and many implicated heads also show similar attention behavior across languages. Together, these results suggest that cross-lingual overlap in agreement circuitry reflects not only shared localization, but also partially shared functional roles.

These findings extend prior demonstrations that cross-lingual circuit overlap can occur by identifying conditions under which that overlap strengthens or weakens: sharing is shaped less by language identity alone than by whether the same grammatical computation must be overtly realized. This is clearest in the contrast between our two recovery metrics: sharing is strongest when the analysis isolates recovery of the inflectional contrast itself, and weaker when it targets broader context-appropriate form recovery. English provides especially strong evidence for this view. It does not behave like a strongly conjugating language overall, but its third-person singular cases become more similar to conjugating languages precisely in the contexts where overt agreement must be computed. This suggests that multilingual models reuse shared circuitry most readily when languages instantiate the same agreement-relevant operation, rather than simply because they belong to the same broad language class.

More broadly, our findings suggest that shared multilingual computation is shaped not only by semantics, but also by whether languages realize the same structural feature in comparable ways. This has implications for multilingual interpretability, as well as for when causal interventions or alignment methods developed in one language should be expected to transfer to others. 

These findings should be interpreted in light of several limitations. Our analysis characterizes the attention-based routing component of agreement rather than the full computational graph: although attention heads enable the inter-token transfer of subject information, position-wise MLPs may play important downstream roles in reading, transforming, or amplifying this signal. Moreover, because we use structured conjugation prompts rather than naturalistic sentences, the identified attention-head signatures may not fully generalize to agreement processing in ordinary language contexts. We also focus on one morphosyntactic phenomenon, on attention-head signatures rather than full circuit reconstruction, and on filtered cases in which the model successfully produces the target form. Extending this analysis to additional grammatical phenomena, model components, and training settings is an important direction for future work. More generally, we hope this work helps shift multilingual mechanistic interpretability from documenting whether internal structure is shared to explaining when grammatical computations recruit shared rather than language-specific mechanisms.

\section*{Acknowledgments}
Part of this work was carried out during the JSALT 2025 workshop as part of the Play-your-Part team. The workshop was supported by discretionary funds from Johns Hopkins University and by the Ministry of Education, Youth and Sports of the Czech Republic through the OP JAK project “Linguistics, Artificial Intelligence and Language and Speech Technologies: From Research to Applications” (project ID CZ.02.01.01/00/23 020/0008518).

This work received funding from the French National Research Agency through grants ANR-16-CONV-0002 (Institute for Language, Communication and the Brain; ILCB) and ANR-20-CE23-0012-01 MIM; the European Union’s Horizon 2020 research and innovation programme under Marie Skłodowska-Curie grant agreement No. 101007666; MCIN/AEI/10.13039/501100011033 under grant PID2024-155948OB-C53; and the Government of Aragon under grant T36 23R. This work was also enabled in part by a gift from the Chan Zuckerberg Initiative Foundation to establish the Kempner Institute for the Study of Natural and Artificial Intelligence.

\bibliography{colm2026_conference}
\bibliographystyle{colm2026_conference}

\appendix
\clearpage
\onecolumn
\section{Task and Experimental Details}
\label{app:setup}

\subsection{Conjugation task and agreement patterns}
\label{app:task}
\label{app:agreement_patterns}

\begin{figure}[H]
  \centering
  \includegraphics[width=0.7\linewidth]{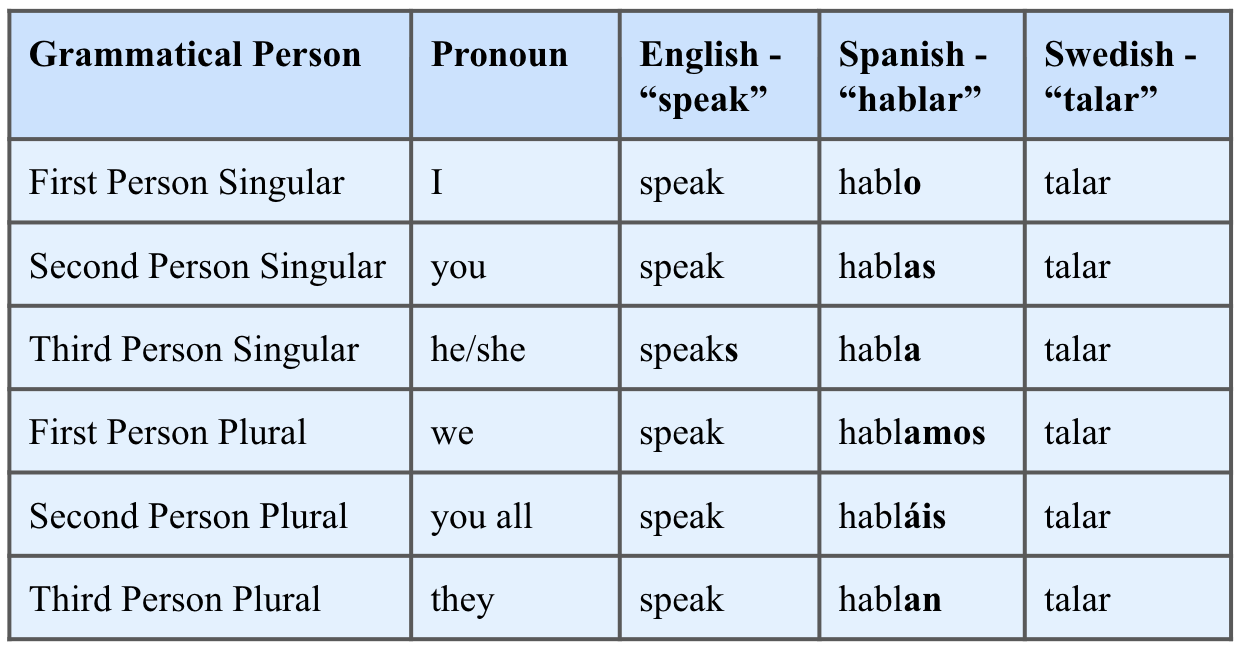}
  \caption{Contrastive examples of verb conjugation across three typologically distinct languages. English only inflects verbs in the third-person singular. Spanish exhibits rich person/number inflection. Swedish does not inflect for person in the tested present-tense contexts.}
  \label{fig:conjugation}
\end{figure}

We study present-tense subject--verb agreement through controlled contrasts in grammatical person and number. Our goal is to isolate when a change in the subject requires a change in the surface form of the verb, and to compare how this computation is implemented across languages. Figure~\ref{fig:conjugation} shows representative examples from three typologically distinct cases: English, which marks agreement only in the third-person singular; Spanish, which exhibits rich person/number inflection; and Swedish, which does not mark person in the tested present-tense contexts.

For reference, Figure~\ref{fig:prompt} lists the full language inventory drawn from our datasets, and Table~\ref{tab:conjlangs} groups the languages retained for analysis by whether the tested contrasts involve overt agreement marking. This organization is important for the main comparisons in the paper, since our cross-lingual analyses distinguish languages with overt agreement from languages where the same contrasts are not morphologically realized.

\subsection{Models used in the main experiments}
\label{app:modelss}

We evaluate the main analyses on a diverse set of open-source multilingual model families. This set gives us variation in architecture, training corpus, and scale while keeping the experimental pipeline fixed across models.

\begin{table}[H]
  \centering
  \caption{Models used in the main experiments.}
  \label{tab:appendix_models_main}
  \begin{tabular}{lll}
    \toprule
    \textbf{Family} & \textbf{Hugging Face Name} & \textbf{Citation} \\
    \midrule
    BLOOM & bigscience/bloom-7b1 & \citep{scao2023bloom} \\
    Gemma & google/gemma-2-9b & \citep{riviere2024gemma2} \\
    Llama & Llama-2-7b-hf & \citep{touvron2023llama2}\\
     & meta-llama/Llama-3.2-3B & \citep{grattafiori2024llama3herdmodels} \\
    Mistral & mistralai/Mistral-7B-v0.1 & \citep{jiang2023mistral7b} \\
    Qwen & Qwen/Qwen2-7B & \citep{yang2024qwen2} \\
    \bottomrule
  \end{tabular}
\end{table}

\FloatBarrier

\subsection{Prompt templates and pronoun inventories}
\label{app:prompts}

For each language, we use a fixed zero-shot prompt template designed to elicit the target present-tense form. The subject pronoun is always stated explicitly, ensuring that the agreement cue is overtly available in every condition. This helps make the cross-lingual comparisons more interpretable: when models differ, the differences are less likely to be driven by whether the subject cue was present in the prompt at all.

Figure~\ref{fig:prompt} displays the full language-specific prompt templates, and the following figure lists the pronoun inventories used for the tested person and number contrasts. Together, these materials define the multilingual prompt space from which the clean/corrupted minimal pairs are constructed.

\begin{figure}[H]
  \centering
  \includegraphics[width=\linewidth]{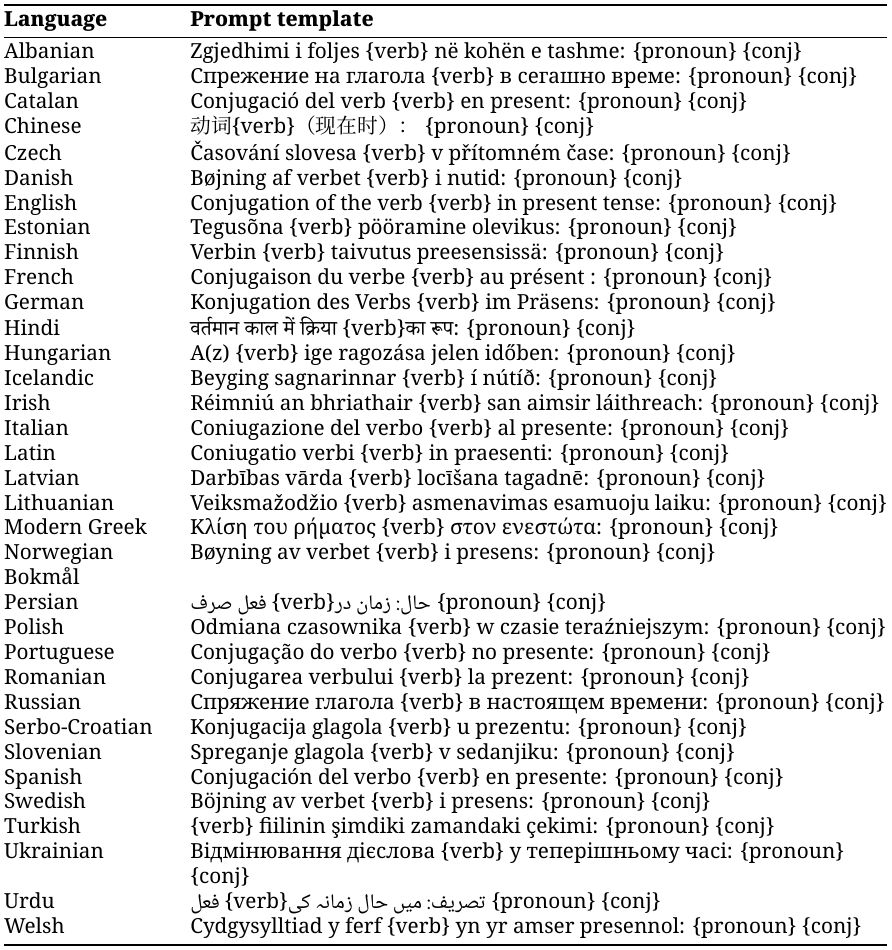}
  \caption{Language-specific prompt templates for all tested languages.}
  \label{fig:prompt}
\end{figure}

\begin{figure}[H]
  \centering
  \includegraphics[width=0.7\linewidth]{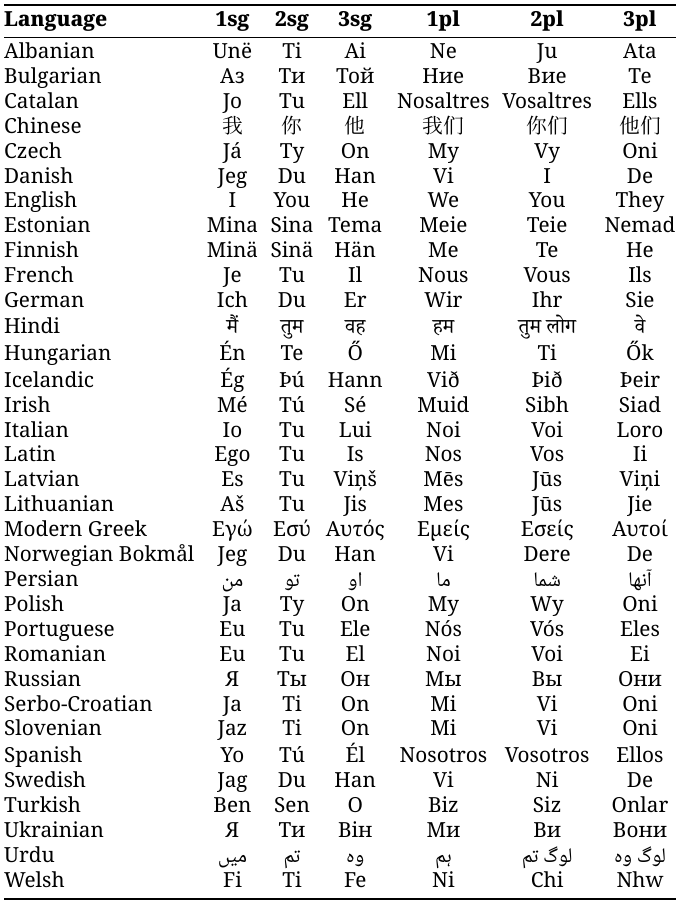}
  \caption{Subject pronouns used in each tested language.}
  \label{fig:pronouns}
\end{figure}

\subsection{Tokenization Filtering}
\label{app:tokenization_examples}

To isolate the agreement contrast as tightly as possible, we retain only cases in which the clean and corrupted conjugated forms differ in the final token and otherwise remain aligned. This model-specific filtering step is crucial for the patching analysis: it ensures that the manipulated contrast is localized to the token being predicted rather than spread across multiple positions in the sequence.

\begin{figure}[H]
  \centering
  \includegraphics[width=0.7\linewidth]{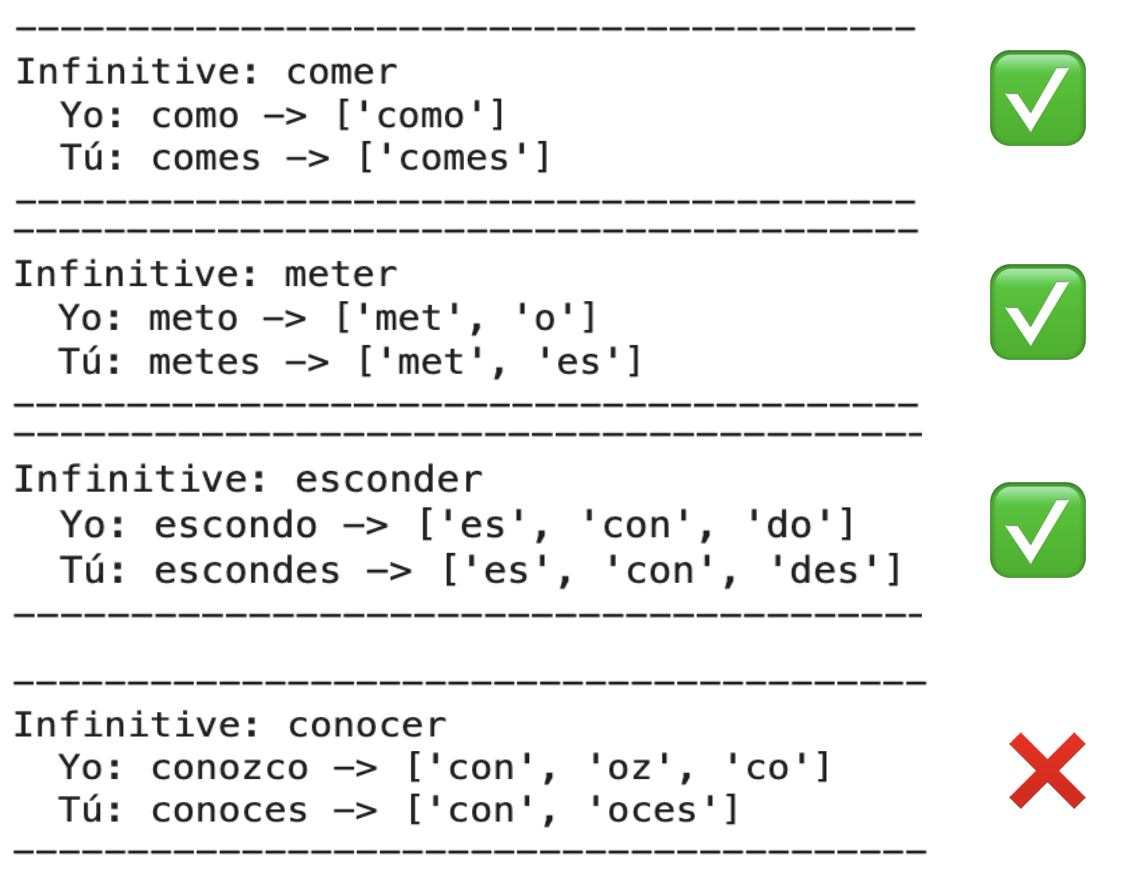}
  \caption{Examples retained and discarded by the tokenization filter.}
  \label{fig:app_tokenization_examples}
\end{figure}

Figure~\ref{fig:app_tokenization_examples} illustrates representative retained and discarded cases. Retained examples preserve the shared prefix and differ only where the agreement contrast should matter most, whereas discarded examples introduce broader tokenization differences that would confound causal interpretation.

\subsection{Languages Tested}
\label{subsec:lang_model}

\subsubsection{Languages In Dataset}
\label{app:languages_tested}

We begin from a broad multilingual inventory assembled from UniMorph and the Chinese Verb Semantic Feature Dataset. Not every language survives the later filtering steps in every model, but this initial inventory gives us wide typological coverage and includes both overtly conjugating and non-conjugating languages.

\begin{table}[H]
\centering
\small
\begin{tabular}{p{0.95\linewidth}}
\toprule
\textbf{Languages} \\
\midrule
Albanian, Bulgarian, Catalan, Chinese, Czech, Danish, English, Estonian, Finnish, French, German, Hindi, Hungarian, Icelandic, Irish, Italian, Latin, Latvian, Lithuanian, Modern Greek, Norwegian Bokm{\aa}l, Persian, Polish, Portuguese, Romanian, Russian, Serbo-Croatian, Slovenian, Spanish, Swedish, Turkish, Ukrainian, Urdu, Welsh \\
\bottomrule
\end{tabular}
\caption{Languages tested across all models (pulled from dataset). Not all of these languages were used in analysis if they did not pass the filtering.}
\label{tab:tested_languages_all_models}
\end{table}

\subsubsection{Languages Used In Analysis}
\label{app:languages_remaining}

After tokenization and behavioral filtering, the analysis set partitions naturally into three groups: languages with overt agreement in the tested contrasts, languages without overt agreement in those contrasts, and English as a middle case. This grouping is the basis for the cross-group comparisons in the main text and appendix.

\begin{table}[H]
\centering
\small
\setlength{\tabcolsep}{6pt}
\renewcommand{\arraystretch}{1.15}
\begin{tabular}{p{0.17\linewidth} p{0.75\linewidth}}
\toprule
\textbf{Group} & \textbf{Languages} \\
\midrule
Conjugating &
Albanian, Bulgarian, Catalan, Czech, Estonian, Finnish, French, German, Hindi, Hungarian, Icelandic, Italian, Latin, Latvian, Lithuanian, Modern Greek, Polish, Portuguese, Romanian, Russian, Serbo-Croatian, Spanish, Turkish, Urdu \\

Nonconjugating &
Chinese, Danish, Norwegian Bokm{\aa}l, Swedish \\

Middle &
English \\
\bottomrule
\end{tabular}
\caption{Languages that survive the filtering and appear in at least one patch result (in either the target-logit or minimal-pair recovery) in at least one model, grouped by agreement type. These are the ones used in analysis.}
\label{tab:conjlangs}
\end{table}

\subsubsection{Language-model coverage}
\label{subsec:lang_modell}

\begin{figure}[H]
  \centering
  \includegraphics[width=0.75\linewidth]{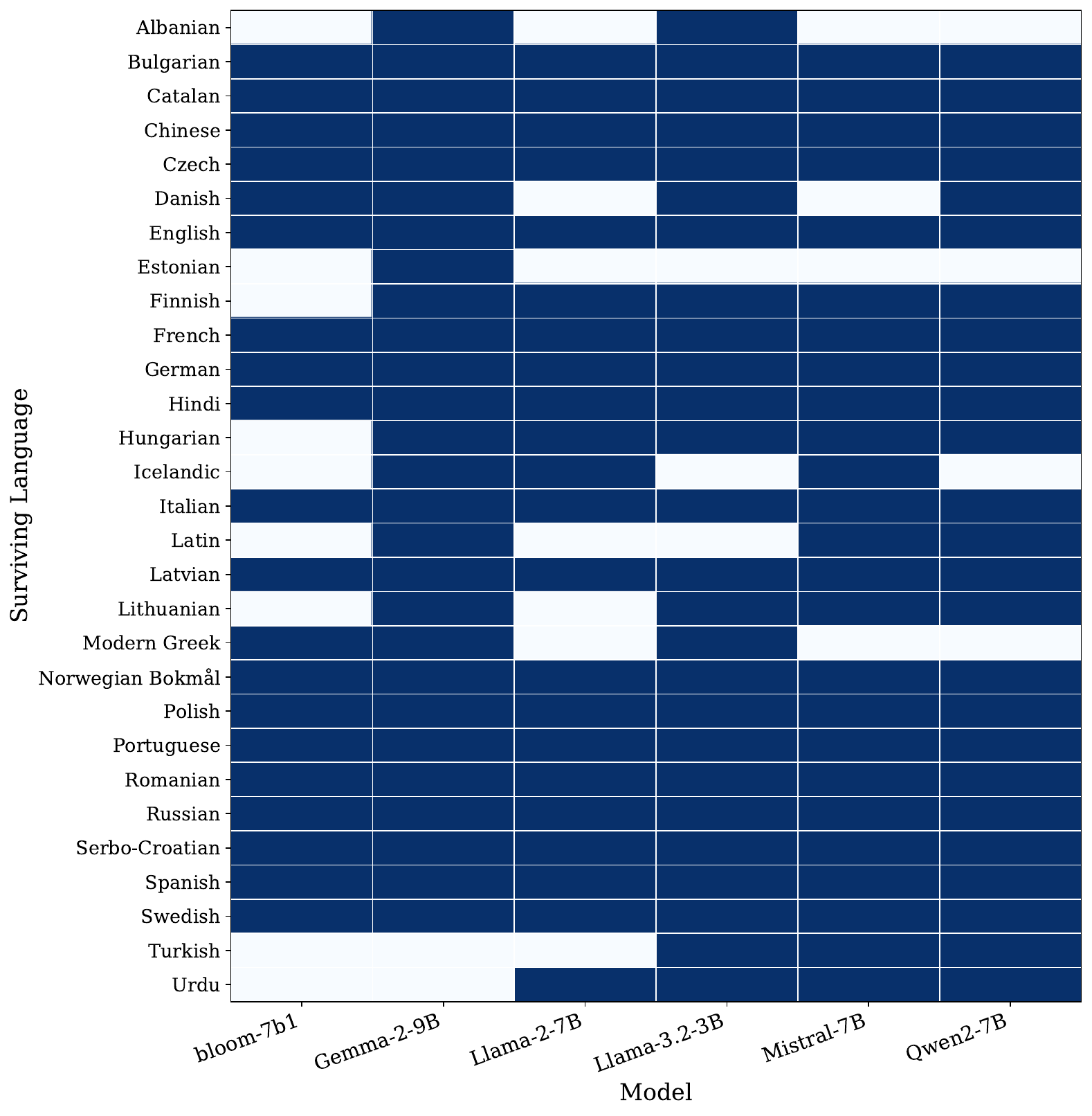}
  \caption{Languages surviving for tested models.}
  \label{fig:appendix_language_coverage}
\end{figure}

Official language support varies across families, but we retain a language--model combination whenever at least 50 prompts in that language remain after tokenization and behavioral filtering. Figure~\ref{fig:appendix_language_coverage} shows that coverage is broad but uneven across models. Gemma, Mistral, and Qwen retain especially wide coverage, whereas BLOOM and the Llama models exhibit more gaps after filtering. Even so, every model contributes a substantial multilingual set, allowing the main cross-lingual comparisons to be matched within model family.

\section{Supplementary Results for Main Claims}
\label{app:supplementary_results}

\subsection{Model-specific and individual language-level cross-lingual similarity among conjugating languages.}
\label{app:heatmaps}

\begin{figure}[H]
    \centering
    \includegraphics[width=\textwidth]{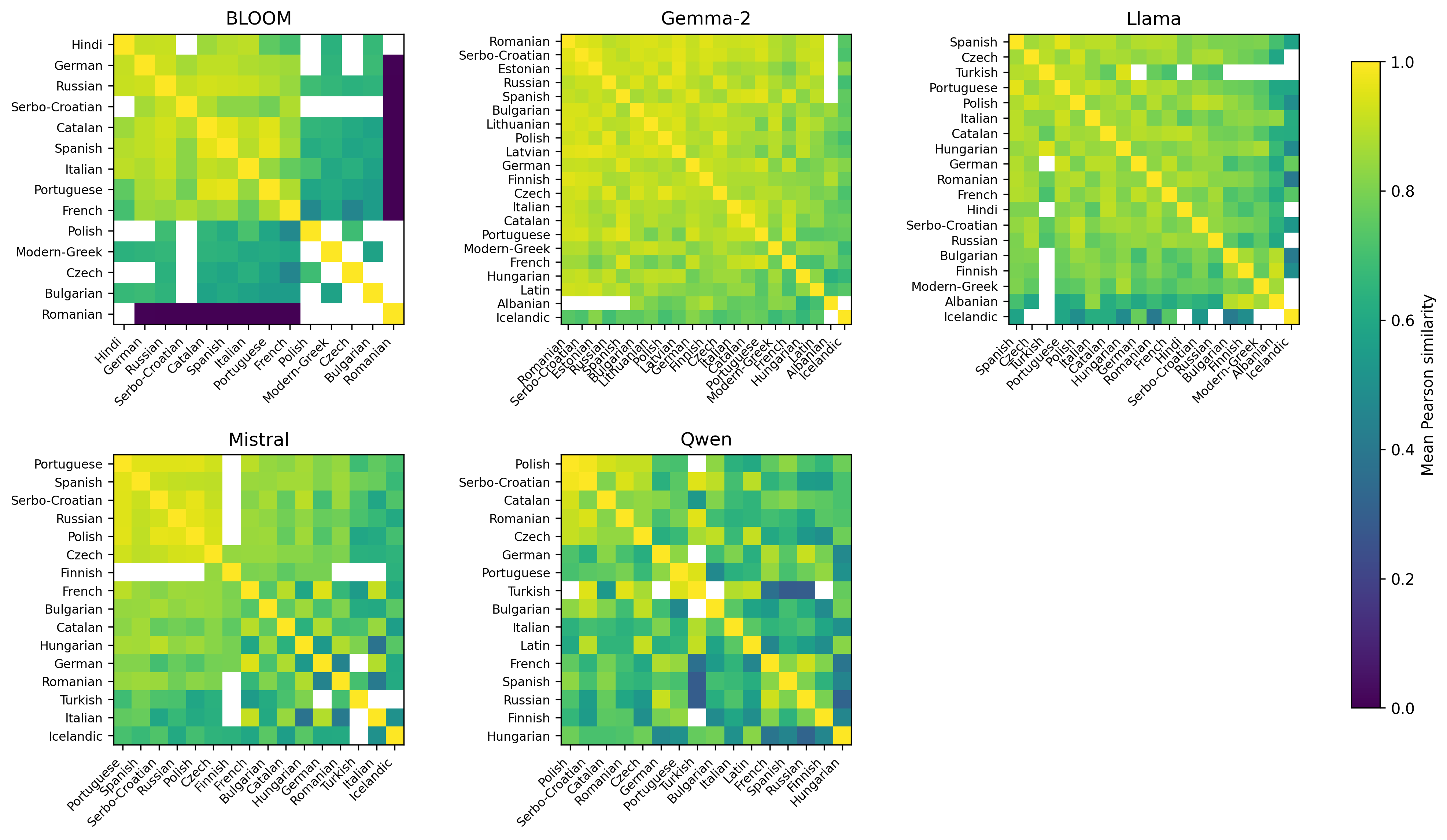}
    \caption{Family-specific language-similarity matrices for conjugating languages under the minimal-pair recovery metric. Each panel shows mean pairwise Pearson similarity between flattened absolute heatmaps, computed at the model-pair level and then averaged within family. White cells denote missing comparisons.}
    \label{fig:family_similarity_matrices}
\end{figure}

Figure~\ref{fig:family_similarity_matrices} expands the main-text result that conjugating languages exhibit positive cross-lingual similarity under the minimal-pair metric. The central pattern is broad, not sparse: within each family, the similarity matrices are dominated by positive values rather than by a few isolated high-similarity pairs. In other words, the effect is not carried by only one language cluster or one especially favorable comparison.

The strength of the pattern does vary by family. Gemma, Mistral, and Qwen show especially dense blocks of positive similarity, while BLOOM and Llama display somewhat more heterogeneity and more missing cells. Importantly, the white cells mark missing comparisons due to coverage constraints rather than evidence against cross-lingual sharing. Overall, these matrices reinforce the claim that shared agreement circuitry among conjugating languages is a broad multilingual phenomenon rather than a narrow family-specific artifact.

\subsection{Similarity by Conjugation Contrast and Patching Direction}
\label{app:direction-split}

To test whether the cross-lingual sharing observed in the main analysis is driven by particular agreement contrasts or patching directions, we disaggregate minimal-pair similarity by conjugation pair and direction. Table~\ref{tab:direction-similarity} reports the mean pairwise Pearson similarity between conjugating languages, averaged across model checkpoints. Cross-lingual similarity remains positive in every condition, indicating that the main pattern is not restricted to a particular patching direction, although the magnitude of similarity varies across contrasts.

\begin{table}[H]
\centering
\small
\begin{tabular}{llrrr}
\toprule
\textbf{Conjugation} & \textbf{Direction} & \textbf{Mean} & \textbf{SD} & \textbf{$n$} \\
\midrule
1sg--2sg & 1sg $\rightarrow$ 2sg & 0.6889 & 0.2950 & 1459 \\
1sg--2sg & 2sg $\rightarrow$ 1sg & 0.6331 & 0.3269 & 1483 \\
1sg--3sg & 1sg $\rightarrow$ 3sg & 0.6045 & 0.3017 & 1476 \\
1sg--3sg & 3sg $\rightarrow$ 1sg & 0.6388 & 0.2887 & 1508 \\
1pl--3pl & 1pl $\rightarrow$ 3pl & 0.5177 & 0.3467 & 1290 \\
1pl--3pl & 3pl $\rightarrow$ 1pl & 0.5040 & 0.3312 & 1320 \\
3sg--3pl & 3pl $\rightarrow$ 3sg & 0.5628 & 0.3247 & 1558 \\
3sg--3pl & 3sg $\rightarrow$ 3pl & 0.5362 & 0.3039 & 1528 \\
\midrule
All & All directions & 0.5876 & 0.3204 & 11622 \\
\bottomrule
\end{tabular}
\caption{Cross-lingual activation-patching similarity among conjugating languages, disaggregated by conjugation contrast and patching direction, where $n$ denotes the number of matched language-pair observations contributing to each condition. Values report mean pairwise Pearson similarity between absolute minimal-pair recovery heatmaps, averaged across model checkpoints. The positive similarity observed in the aggregate analysis persists across all individual directions, although its magnitude varies across contrasts.}
\label{tab:direction-similarity}
\end{table}

\subsection{Model-specific cross-group comparisons}
\label{app:cross_group_model_details}

\begin{figure}[H]
    \centering
    \includegraphics[width=\textwidth]{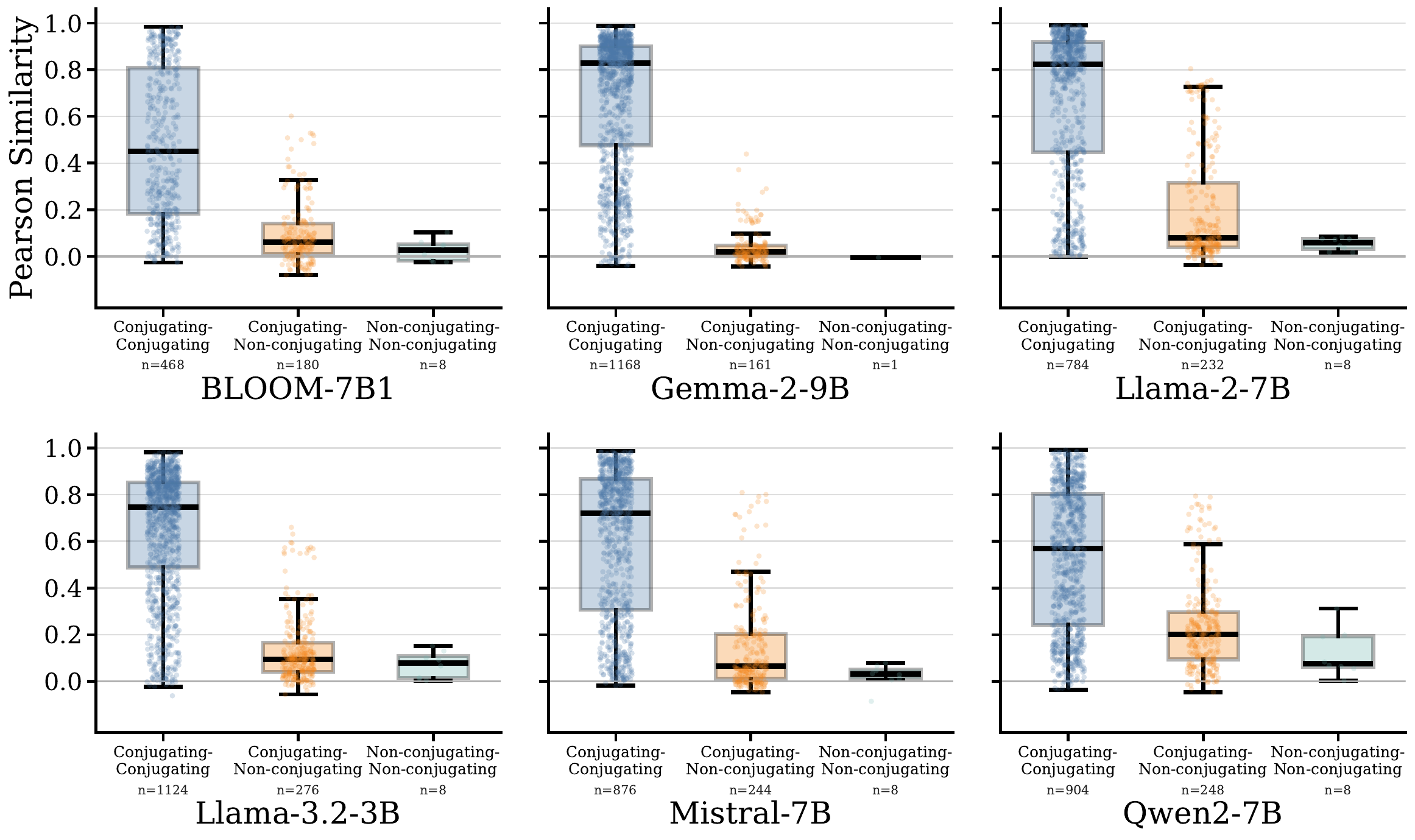}
    \caption{Model-specific cross-group similarity distributions. Within each panel, conjugating--conjugating comparisons are consistently more similar than conjugating--non-conjugating comparisons, although the size of the gap varies by model family.}
    \label{fig:cross_group_model_small_multiples}
\end{figure}

Figure~\ref{fig:cross_group_model_small_multiples} breaks out the main cross-group comparison by model family. The same ordering appears in every family: conjugating--conjugating comparisons are highest, conjugating--non-conjugating comparisons are lower, and non-conjugating--non-conjugating comparisons remain lowest overall. This consistency matters because it shows that the grammatical separation highlighted in the main text is not being driven by a single architecture. Taken together, these panels show that the cross-group separation is stable across model families, even though the magnitude of the gap varies.

\FloatBarrier

\subsection{Additional analyses of English's mixed agreement profile}
\label{app:english_mixed_profile}
\label{app:english_model_deltas}

\begin{figure}[H]
    \centering
    \includegraphics[width=0.7\linewidth]{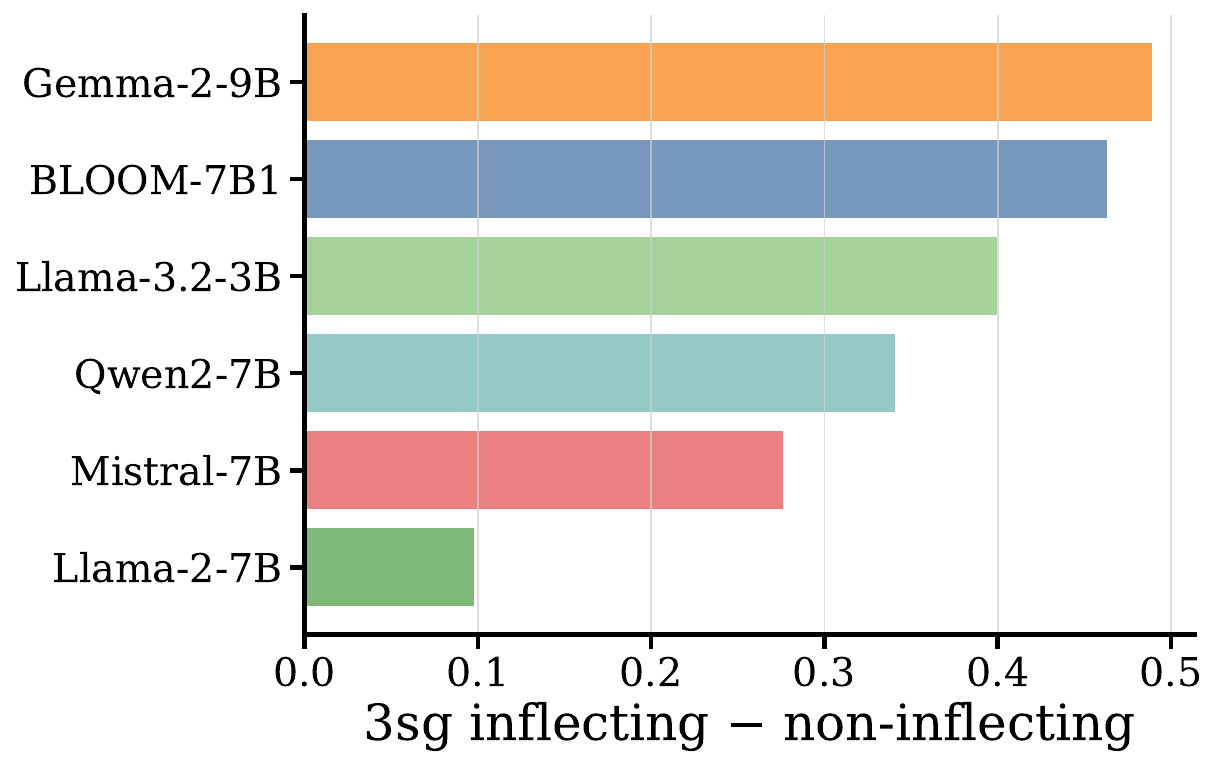}
    \caption{English shift toward conjugating-language circuitry, by model.}
    \label{fig:english_model_deltass}
\end{figure}

Figure~\ref{fig:english_model_deltass} provides a model-by-model view of the English effect from the main text. The result is directionally consistent across all six models: English is always more similar to conjugating-language circuitry in the 3sg-inflecting conditions than in the non-inflecting conditions. What changes across models is the size of the shift, not its sign.

The largest shifts appear in Gemma-2-9B and BLOOM-7B1, with Llama-3.2-3B and Qwen2-7B also showing clear positive differences. Mistral-7B and Llama-2-7B exhibit smaller but still positive shifts. This model-level consistency strengthens the interpretation that English behaves as a true bridge case: it aligns more closely with conjugating languages precisely in the contexts where English itself overtly realizes agreement.

\FloatBarrier

\FloatBarrier

\section{Robustness to Similarity Metric}
\label{app:robustness_checks_metric}

\begin{figure}[H]
    \centering
    \includegraphics[width=0.7\linewidth]{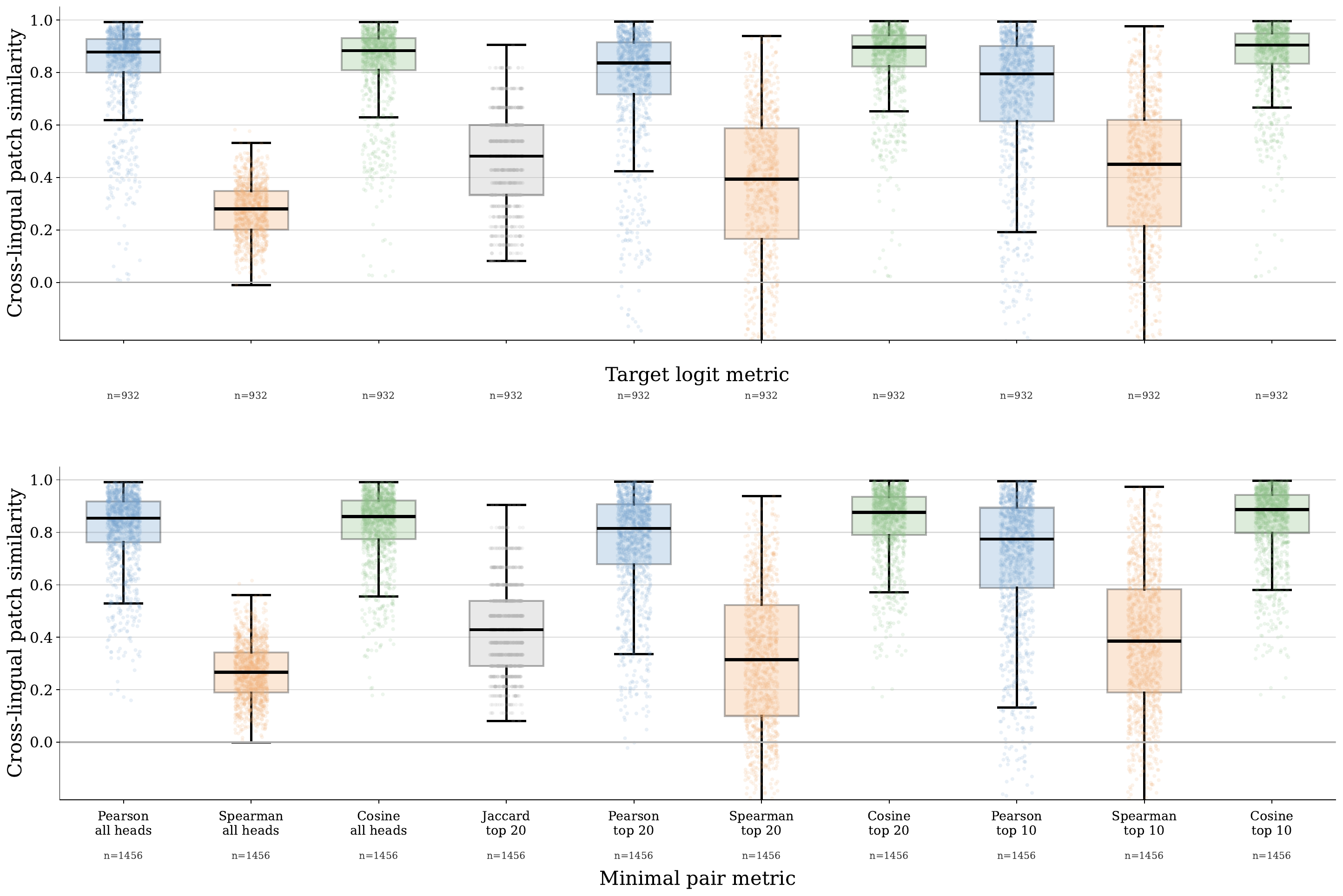}
    \caption{Cross-lingual similarity of conjugating languages under alternative similarity metrics and head-selection schemes. Pearson and cosine recover the main pattern most consistently, whereas Spearman is visibly noisier.}
    \label{fig:conj_model_deltas}
\end{figure}

Figure~\ref{fig:conj_model_deltas} compares the cross-lingual similarity result across several alternative similarity metrics and head-selection schemes. The broad conclusion is that the main pattern is robust, but some metrics capture it more cleanly than others. Pearson and cosine similarity reproduce the effect most stably across both all-head and top-$k$ variants, while Spearman is visibly noisier and less discriminating. Jaccard overlap on the top-20 heads captures some shared structure, but compresses the distribution and loses graded information relative to correlation-based measures.

For this reason, the main text focuses on Pearson correlation over flattened absolute heatmaps. It preserves the continuous structure of the patching signature and yields the clearest separation between stronger and weaker cross-lingual overlap. The robustness figure therefore supports the substantive claim while also motivating the specific similarity metric used in the paper.

\FloatBarrier

\section{Auxiliary Analyses}
\label{app:robustness_checks}

For the auxiliary analyses, we work with a smaller language subset (Table \ref{tab:agreement_patternsss}) chosen to preserve typological diversity while making the expanded model comparisons computationally tractable. This subset includes richly conjugating languages, English as a mixed case, and non-conjugating comparison languages.

\begin{table}[H]
  \centering
  \begin{tabular}{p{11cm}}
    \toprule
    \textbf{Languages} \\
    \midrule
    Spanish, Portuguese, Italian, French, Catalan, Czech, Finnish, Russian, Hungarian, Polish, Serbo-Croatian, German, English, Swedish, Chinese, Mongolian \\
    \bottomrule
  \end{tabular}
  \caption{Languages included in the auxiliary experiments.}
  \label{tab:agreement_patternsss}
\end{table}

\subsection{Model size and language accuracy}

\begin{figure}[H]
    \centering
    \begin{subfigure}[t]{0.53\textwidth}
        \centering
        \includegraphics[width=\linewidth]{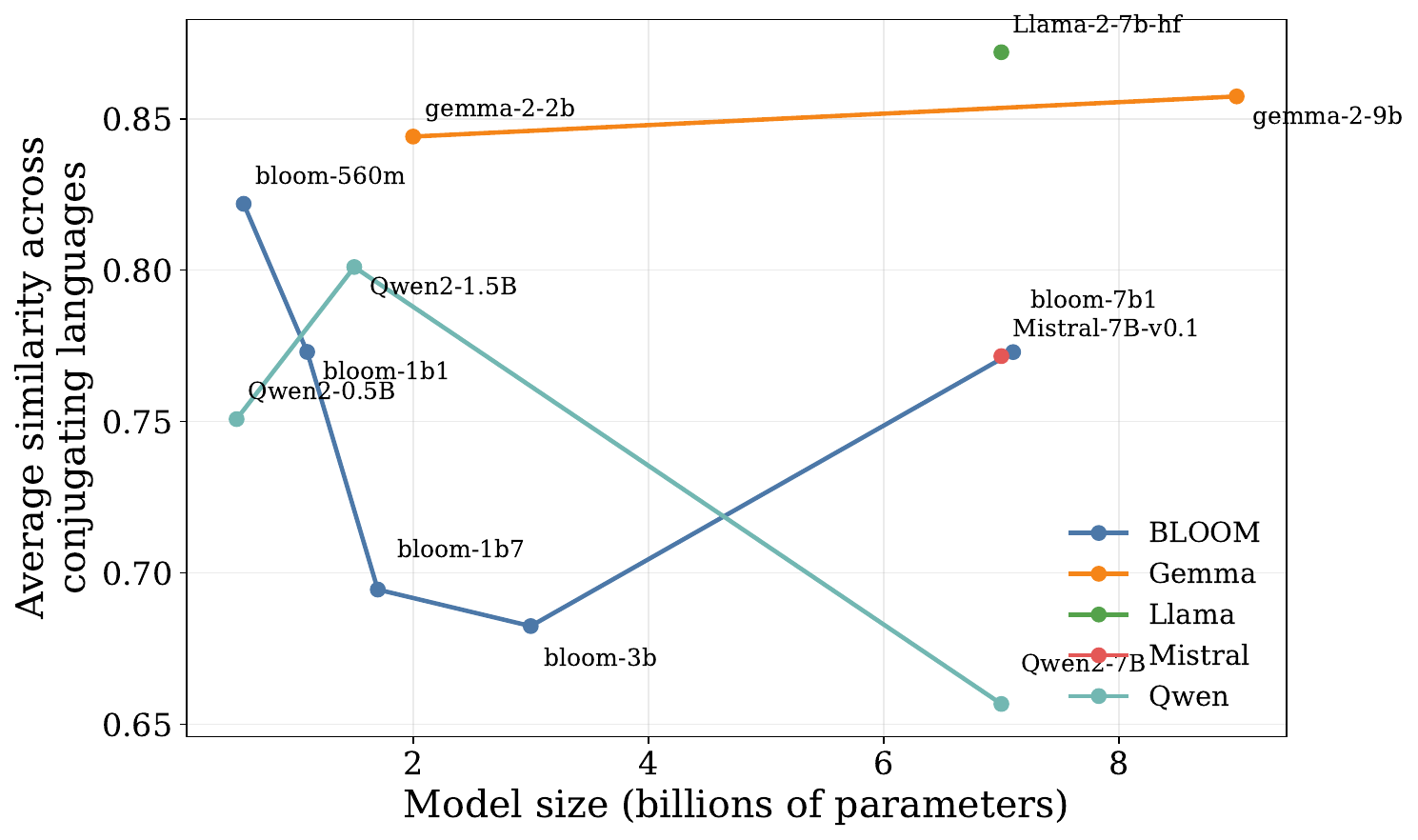}
        \caption{Cross-lingual similarity as a function of model scale.}
        \label{fig:app_model_scale_similarity}
    \end{subfigure}
    \hfill
    \begin{subfigure}[t]{0.44\textwidth}
        \centering
        \includegraphics[width=\linewidth]{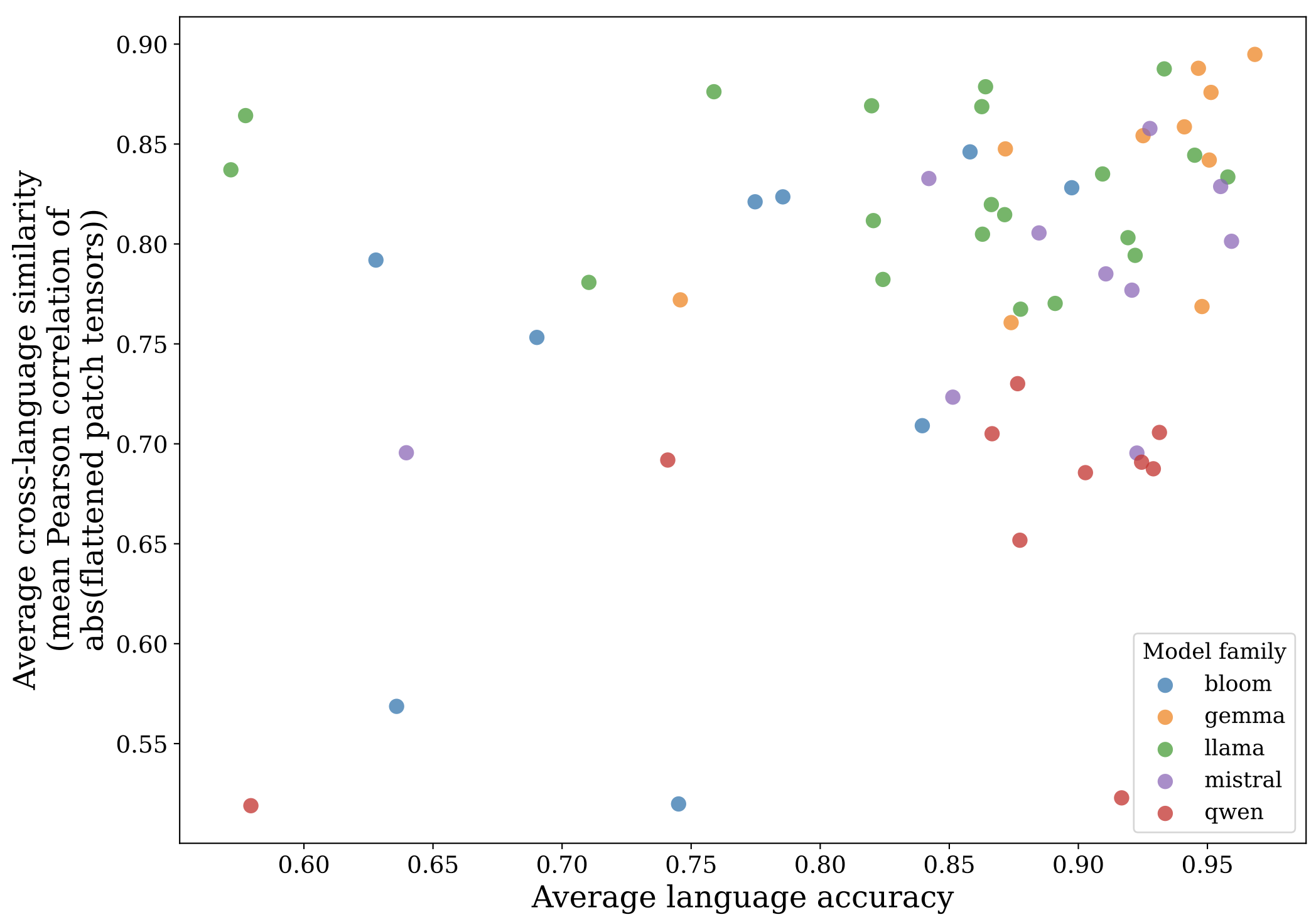}
        \caption{Cross-lingual similarity versus language-specific task accuracy.}
        \label{fig:app_accuracy_vs_similarity}
    \end{subfigure}
    \caption{Robustness checks for model scale and task accuracy.}
    \label{fig:app_scale_accuracy}
\end{figure}

These analyses test whether the cross-lingual similarity pattern could be reduced to either model scale or language-specific task performance. Figure~\ref{fig:app_scale_accuracy} suggests that neither account is sufficient on its own.

The model-scale analysis does not show a single monotonic relationship between parameter count and cross-lingual similarity. Within some families, similarity rises with scale, but in others it flattens or changes only modestly. Likewise, the accuracy analysis shows that better language-specific task performance does not mechanically imply stronger cross-lingual circuit similarity. Languages and models with similar accuracy can still differ noticeably in cross-language overlap. These results support the view that the main similarity pattern reflects something more specific than generic model quality or scale.

\begin{table}[H]
  \centering
  \caption{Models used in the model-scale and accuracy robustness analyses.}
  \label{tab:robustness_models}
  \begin{tabular}{lll}
    \toprule
    \textbf{Family} & \textbf{Hugging Face Name} & \textbf{Citation} \\
    \midrule
    BLOOM & bigscience/bloom-560m & \citep{scao2023bloom} \\
    BLOOM & bigscience/bloom-1b1 & \citep{scao2023bloom} \\
    BLOOM & bigscience/bloom-1b7 & \citep{scao2023bloom} \\
    BLOOM & bigscience/bloom-3b & \citep{scao2023bloom} \\
    BLOOM & bigscience/bloom-7b1 & \citep{scao2023bloom} \\
    Gemma & google/gemma-2-9b & \citep{riviere2024gemma2} \\
    Gemma & google/gemma-2-2b & \citep{riviere2024gemma2} \\
    Llama & Llama-2-7b-hf & \citep{touvron2023llama2}\\
    Mistral & mistralai/Mistral-7B-v0.1 & \citep{jiang2023mistral7b} \\
    Qwen & Qwen/Qwen2-0.5B & \citep{yang2024qwen2} \\
    Qwen & Qwen/Qwen2-1.5B & \citep{yang2024qwen2} \\
    Qwen & Qwen/Qwen2-7B & \citep{yang2024qwen2} \\
    \bottomrule
  \end{tabular}
\end{table}

\FloatBarrier

\subsection{Instruction tuning}

\begin{figure}[H]
    \centering
    \includegraphics[width=0.8\linewidth]{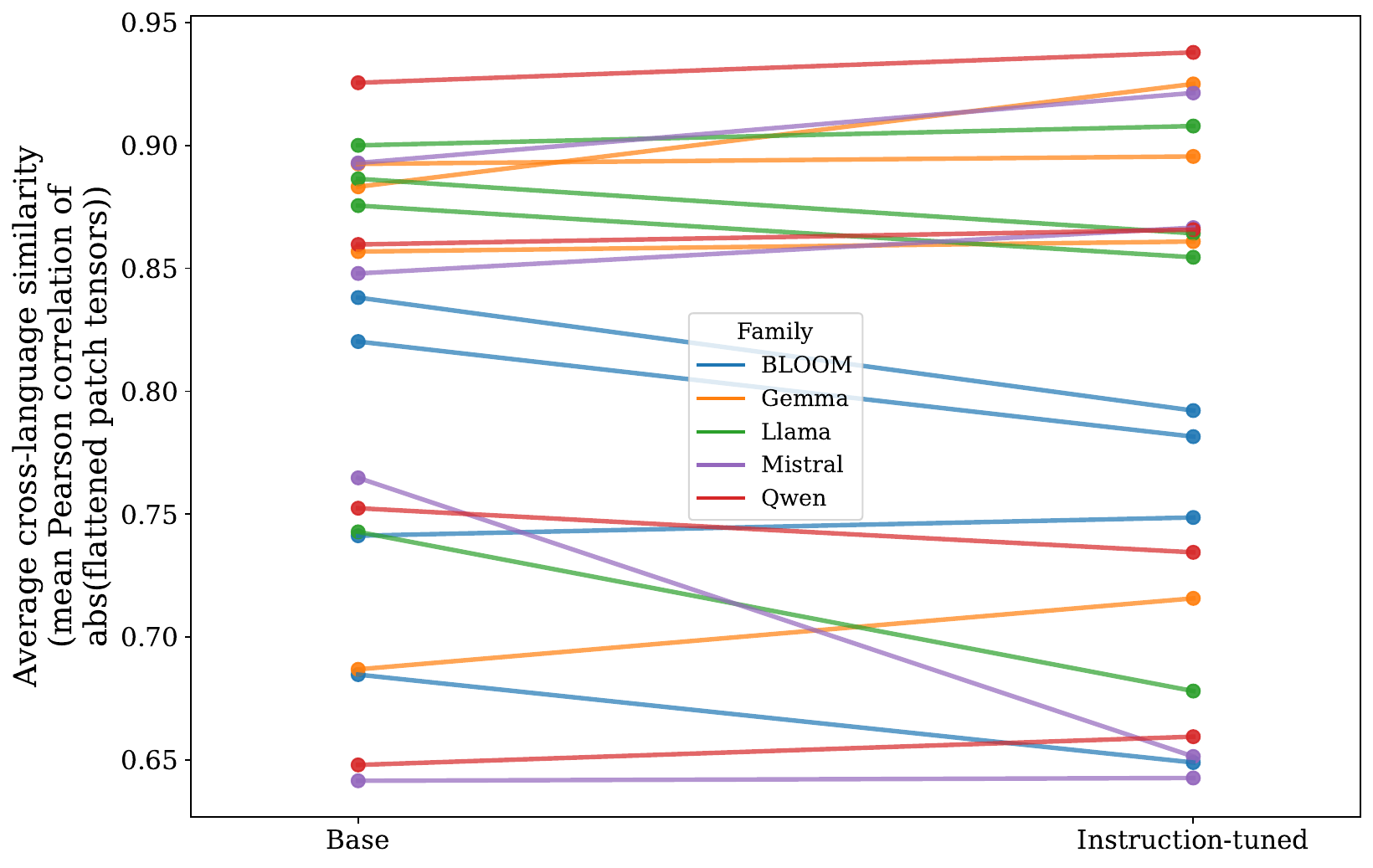}
    \caption{Instruction tuning leaves the overall similarity structure largely unchanged.}
    \label{fig:english_similarity_boxstrip}
\end{figure}

We also test whether the recovered similarity structure changes coherently after instruction tuning. Figure~\ref{fig:english_similarity_boxstrip} shows that the answer is largely no: instruction tuning does not introduce a consistent upward or downward shift in cross-lingual similarity across families.

Any changes that do appear are modest and mixed in direction. This suggests that the circuit-level cross-lingual structure relevant to agreement is already present in the base models and is not substantially reorganized by instruction tuning. In other words, the sharing pattern documented in the main text appears to be relatively stable under this form of post-training.

\begin{table}[H]
  \centering
  \small
  \begin{tabular}{p{0.25\textwidth} p{0.32\textwidth} p{0.33\textwidth}}
    \toprule
    \textbf{Base} & \textbf{Instruction-tuned} & \textbf{Citation} \\
    \midrule
    bloom-1b7       & bloomz-1b7              & \citep{scao2023bloom, muennighoff-etal-2023-crosslingual} \\
    
    gemma-2-2b      & gemma-2-2b-it           & \citep{riviere2024gemma2} \\
    Llama-2-7b-hf   & Llama-2-7b-chat-hf      & \citep{touvron2023llama2} \\
    Mistral-7B-v0.1 & Mistral-7B-Instruct-v0.1 & \citep{jiang2023mistral7b} \\
    Qwen2-1.5B      & Qwen2-1.5B-Instruct     & \citep{yang2024qwen2} \\
    \bottomrule
  \end{tabular}
  \caption{Base and instruction-tuned models used in the instruction-tuning comparison.}
  \label{tab:instruction_tuning_models}
\end{table}

\FloatBarrier

\subsection{Analyses of pre-training progression}
\label{app:checkpoint_progression}

\begin{figure}[H]
    \centering
    \begin{subfigure}[t]{0.48\textwidth}
        \centering
        \includegraphics[width=\linewidth]{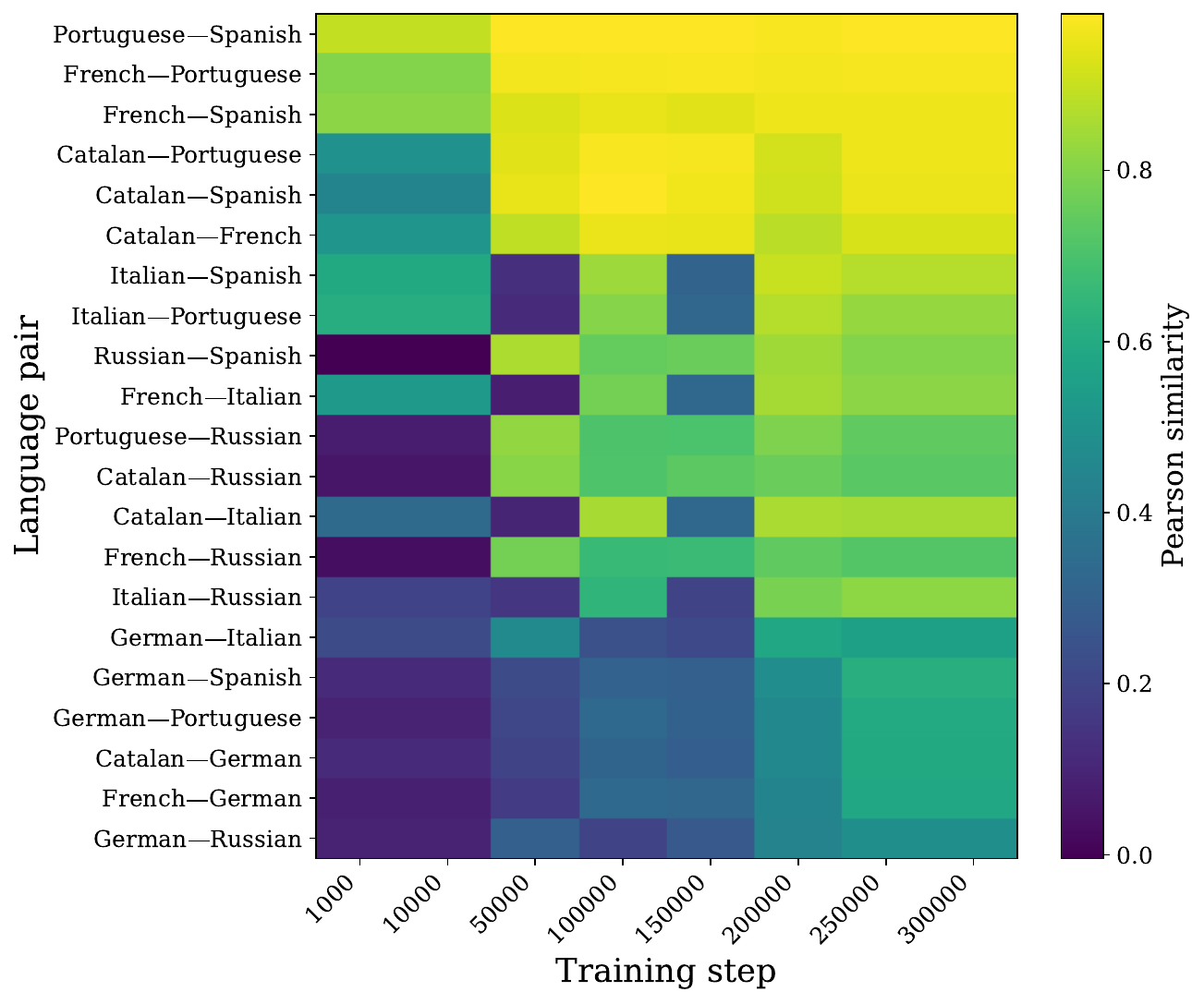}
        \caption{Similarity by language pair and checkpoint in the minimal-pair metric.}
        \label{fig:app_checkpoint_heatmap_closed}
    \end{subfigure}
    \hfill
    \begin{subfigure}[t]{0.48\textwidth}
        \centering
        \includegraphics[width=\linewidth]{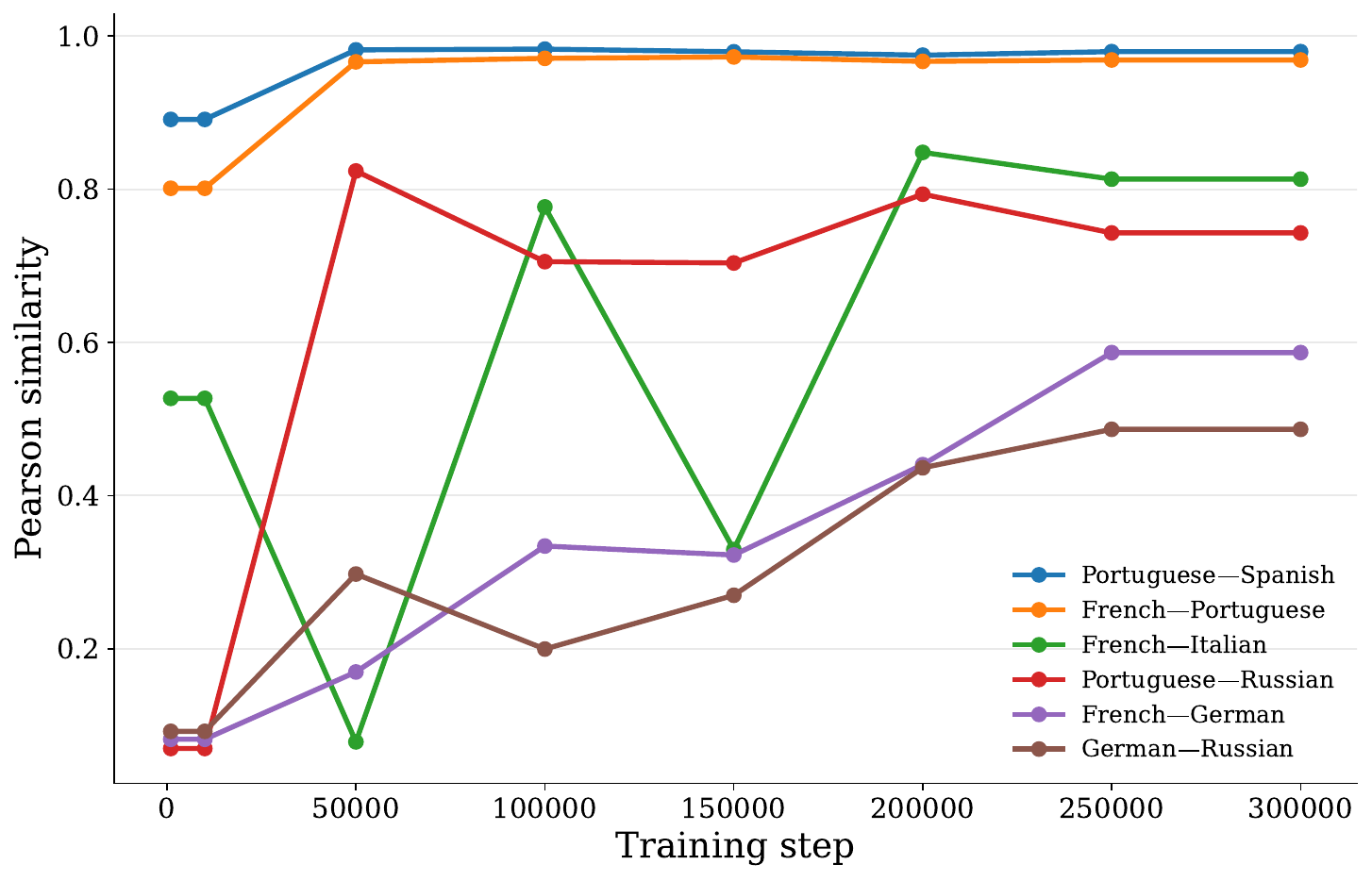}
        \caption{Representative language-pair trajectories in the minimal-pair metric.}
        \label{fig:app_checkpoint_trajectories_closed}
    \end{subfigure}

    \vspace{0.8em}

    \begin{subfigure}[t]{0.62\textwidth}
        \centering
        \includegraphics[width=\linewidth]{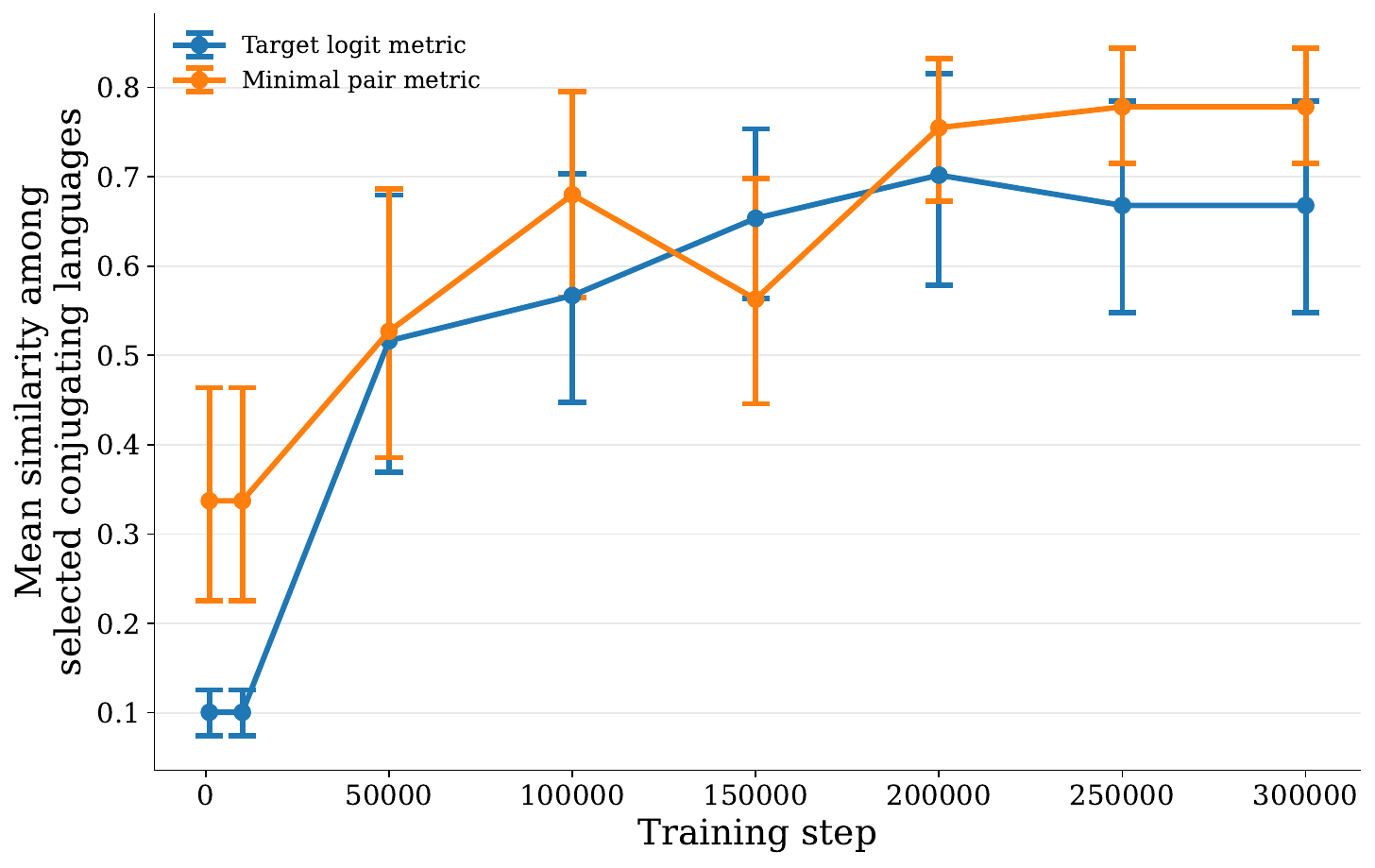}
        \caption{Comparison of target-logit and minimal-pair similarity trajectories.}
        \label{fig:app_checkpoint_open_closed}
    \end{subfigure}
    \caption{Supplementary views of the BLOOM checkpoint analysis. Across both metrics, cross-lingual similarity tends to increase over training, although the minimal-pair metric is noisier and more contrast-specific.}
    \label{fig:app_checkpoint_progression}
\end{figure}

As a training-dynamics case study, we examine intermediate BLOOM checkpoints to ask whether cross-lingual similarity in conjugation circuitry emerges gradually over pretraining. Figure~\ref{fig:app_checkpoint_progression} shows a clear overall trend: across both recovery metrics, similarity tends to increase over training, although the minimal-pair signal is noisier and more contrast-specific than the target-logit signal.

The language-pair heatmap and representative trajectories show that individual pairs do not all move identically. Some pairs rise sharply early and then plateau, while others increase more gradually. But at the aggregate level, the trajectory is upward, indicating that shared cross-lingual agreement circuitry is not fully present from the start and instead becomes more pronounced over pretraining. The comparison between the two metrics further mirrors the main-text story: the target-logit metric produces a smoother developmental trajectory, whereas the minimal-pair metric captures a more selective and variable inflection-specific signal.

\begin{table}[H]
  \centering
  \small
  \begin{tabular}{ll}
    \toprule
    \textbf{Base model} & \texttt{bigscience/bloom-1b7} \\
    \textbf{Checkpoint repo} & \texttt{bigscience/bloom-1b7-intermediate} \\
    \textbf{Steps} & 1k, 10k, 50k, 100k, 150k, 200k, 250k, 300k \\
    \bottomrule
  \end{tabular}
  \caption{Base model and intermediate checkpoints used in the BLOOM progression analysis \citep{scao2023bloom}.}
  \label{tab:checkpoint_models}
\end{table}

\FloatBarrier

\label{sec:appendix}


\end{document}